\PassOptionsToPackage{pagenumbers}{cvpr}

\def\includeAppendix{1}

\documentclass[10pt,twocolumn,letterpaper]{article}

\usepackage{cvpr}              

\usepackage[accsupp]{axessibility}

\usepackage{latexsym}

\let\citet\cite

\usepackage[utf8]{inputenc}
\usepackage[T1]{fontenc}

\usepackage{amsfonts,amsmath,amssymb}       

\usepackage{mathtools}

\usepackage{newtxmath}

\usepackage[mathscr]{euscript}

\usepackage{relsize}       

\usepackage[hyphens]{url}

\usepackage{array}

\usepackage{booktabs}

\usepackage{placeins}
\usepackage{listings}

\usepackage[pagebackref,breaklinks=true,colorlinks]{hyperref}

\usepackage[capitalize]{cleveref}
\crefname{section}{Sec.}{Secs.}
\Crefname{section}{Section}{Sections}
\Crefname{table}{Table}{Tables}
\crefname{table}{Tab.}{Tabs.}

\usepackage{xcolor}

\definecolor{codegreen}{rgb}{0,0.5,0}
\definecolor{codegray}{rgb}{0.5,0.5,0.5}
\definecolor{codepurple}{rgb}{0.58,0,0.82}
\definecolor{backcolour}{rgb}{0.9141,0.9141,0.9414} 

\lstdefinestyle{mystyle}{
  backgroundcolor=\color{backcolour}, commentstyle=\color{codegreen},
  keywordstyle=\color{magenta},
  numberstyle=\tiny\color{codegray},
  stringstyle=\color{codepurple},
  basicstyle=\ttfamily\footnotesize,
breakatwhitespace=false,         
  breaklines=true,                 
  captionpos=b,                    
  keepspaces=true,                 
  numbers=left,                    
  numbersep=1pt,                  
  showspaces=false,                
  showstringspaces=false,
  showtabs=false,                  
  tabsize=2
}

\let\tablefont=\footnotesize

\usepackage{tikz}

\def\transpose{^{\textup{T}}}
\let\mat\mathbf
\def\Rset{\mathbb{R}}

\let\cconv*

\let\hprod\circ

\def\flip{\mathop{\textup{inv}}}
\def\DFT{\mathop{\mathscr{F}}}
\def\IDFT{\mathop{\mathscr{F}^{-1}\!}}

\newcommand\matK{\mat{K}}
\newcommand\matM{\mat{M}}

\newcommand\symA{a}
\newcommand\setA{\mathrm{A}}
\newcommand\matKA{\matK(\setA)}

\newcommand\symB{b}
\newcommand\setB{\mathrm{B}}
\newcommand\matKB{\matK(\setB)}

\makeatletter
\newcommand\vecA[1][]{\def\foo{#1}\ifx\foo\@empty\else\edef\foo{^{(\foo)}}\fi\mat{\symA}\foo}
\newcommand\vecB[1][]{\def\foo{#1}\ifx\foo\@empty\else\edef\foo{^{(\foo)}}\fi\mat{\symB}\foo}
\makeatother

\newcommand\matC{\mat{C}}
\newcommand\matCAB{\matC(\setA, \setB)}

\newcommand\sumsym{v}

\newcommand\sumvec{\mathop{\textup{sumvec}}}

\newcommand\Rsum{\regfunc_{\textup{sum}}}

\newcommand\Roff{\regfunc_{\textup{off}}}
\let\Rbarlow\Roff

\newcommand\Rvar{\regfunc_{\textup{var}}}
\let\Rcov\Rbarlow

\newcommand\lossfunc{L}

\newcommand\regfunc{R}

\newcommand\covsum{Proposed}
\newcommand\xsum{Proposed}

\ifdefined\appendixName
\else
  \usepackage{xspace}
\def\appendixName{Supplementary Material\xspace}
\fi

\ifdefined\includeAppendix
  \let\crefappendix\cref  
\else
  \def\appendix{\end{document}}
  \def\crefappendix#1{\appendixName}
\fi

\ifdefined\separateReferences
\usepackage{multibib}
  \newcites{appendix}{Additional References Cited in \appendixName}
\else \let\citeappendix=\cite
  \let\citetappendix=\citet
  \makeatletter
  \let\bibliographyappendix=\@gobble
  \let\bibliographystyleappendix=\@gobble
  \makeatother
\fi

\def\cvprPaperID{8591} \def\confName{CVPR}
\def\confYear{2023}

\title{Learning Decorrelated Representations Efficiently\\ Using Fast Fourier Transform}

\author{Yutaro Shigeto\thanks{Equal contribution.}\quad
  Masashi Shimbo${}^{*}$\quad
  Yuya Yoshikawa\quad
  Akikazu Takeuchi\\
  \texttt{\{shigeto,shimbo,yoshikawa,takeuchi\}@stair.center}\\
  STAIR Lab, Chiba Institute of Technology, Narashino, Chiba, Japan}

\begin{document}

\maketitle

\begin{abstract}
Barlow Twins and VICReg are self-supervised representation learning models that use regularizers to decorrelate features.
Although these models are as effective as conventional representation learning models, their training can be computationally demanding if the dimension $d$ of the projected embeddings is high.
As the regularizers are defined in terms of individual elements of a cross-correlation or covariance matrix, computing the loss for $n$ samples takes $O(n d^2)$ time.
In this paper, we propose a relaxed decorrelating regularizer that can be computed in $O(n d\log d)$ time by Fast Fourier Transform.
We also propose an inexpensive technique to mitigate undesirable local minima that develop with the relaxation.
The proposed regularizer exhibits accuracy comparable to that of existing regularizers in downstream tasks, whereas their training requires less memory and is faster for large $d$.
The source code is available.\footnote{\url{https://github.com/yutaro-s/scalable-decorrelation-ssl.git}}
\end{abstract}

\section{Introduction}
\label{sec:introduction}

Self-supervised learning (SSL) of representations \cite{Oord2018,Chen2020,Wang2020,He2020,chen2020big,Grill2020,chen2021exploring,Zbontar2021,Bardes2022,ermolov2021whitening,Zhang2022} has become an integral part of deep learning applications in computer vision. Most SSL models for visual representations employ a multi-view, Siamese network architecture.
First, an input image is altered by different transformations that preserve its original semantics. The two augmented examples are then fed to a neural network (or in some cases, two different networks)
that consists of a backbone network cascaded with a small projection network (usually a multi-layer perceptron)
to produce ``twin'' projected embeddings, or views, of the original.
Finally, the network weights are trained so that the twin embeddings (referred to as a ``positive pair'') are similar, reflecting the fact that they represent the same original image.
After training, the projection network is discarded, while the backbone network is reused for downstream tasks;
the idea is that the pretrained backbone should produce generic representations of images that also benefit downstream tasks.

One major issue in SSL is to sidestep \emph{collapsed embeddings}, or the presence of trivial solutions such that all examples are projected to a single embedding vector.
Contrastive approaches \cite{Oord2018,Chen2020,Wang2020,He2020,chen2020big}
eliminate such solutions by a loss term that repels embeddings of different original images (known as ``negative pairs'') from each other.
Considering all possible negative pairs is infeasible, and negative sampling is usually performed.

Recent studies have explored non-contrastive SSL models. Among these models,
Barlow Twins \cite{Zbontar2021} and VICReg \cite{Bardes2022}
use loss functions to penalize features (components of embedding vectors) with a small variance, which are characteristic of collapsed embeddings.
Also notable in these loss functions are regularization terms for feature decorrelation.
They reduce redundancy in the learned features and make Barlow Twins and VICReg perform as well as contrastive models.
However, since the regularizers are defined in terms of elements in a sample cross-correlation or covariance matrix,
they require $O(n d^2)$ time to compute, where $n$ is the number of examples in a batch, and $d$ is the dimensionality of the projected embeddings.
This is problematic, as an increased $d$
has been reported to improve the performance of both Barlow Twins and VICReg \cite{Zbontar2021,Bardes2022}.

\paragraph{Contributions.}

We propose a relaxed decorrelating regularizer to address the inefficiency of Barlow Twins and VICReg.
The proposed regularizer does not require the explicit calculation of cross-correlation or covariance matrices
and can be computed in $O(n d \log d)$ time by means of Fast Fourier Transform (FFT) \cite{Cooley1965fft}.
Although undesirable local minima develop as a result of relaxation, they can be mitigated by feature permutation;
we give an account of why this simple technique works.
The SSL models using the proposed regularizer achieve competitive performance with Barlow Twins and VICReg in downstream tasks,
with substantially less computation time for large $d$.
With $d=8192$, training is 1.2 (with a ResNet-50 backbone) or 2.2 (with a lightweight ResNet-18 backbone) times as fast as Barlow Twins.
The proposed method also reduces memory consumption, which allows for larger batch size.

\paragraph{Notation.}

We use zero-based indexing for vector and matrix components unless stated otherwise;
thus, for a vector $\mat{x} \in \Rset^d$, the component index ranges from $0$ to $d-1$.
We use $[\mat{x}]_i$ to denote the $i$th component of vector $\mat{x}$,
and $[\mat{M}]_{i j}$ to denote the $(i,j)$th element of matrix $\mat{M}$.
For a complex vector $\mat{c}$, $\overline{\mat{c}}$ denotes its componentwise complex conjugate.
For vectors $\mat{x}$ and $\mat{y}$, $\mat{x} \hprod \mat{y}$ denotes their componentwise product.

\section{Related Work}
\label{sec:related}

\paragraph{Contrastive SSL.}

Contrastive SSL uses positive and negative pairs of augmented samples~\cite{Oord2018,Chen2020,Wang2020,He2020,chen2020big,dwibedi2021little,Chen2021,tian2020contrastive}.
The commonly used InfoNCE loss~\cite{Oord2018} consists of an alignment term, which maximizes the similarity between positive pairs,
and a uniformity term, which minimizes the similarity between negative pairs~\cite{Wang2020}.
SimCLR~\cite{Chen2020} is a state-of-the-art contrastive SSL method. However, to obtain effective representations, SimCLR requires a large number of negative pairs \cite{Chen2020},
or, in other words, a large batch (or memory bank) size $n$.
This can be a computational bottleneck, 
as the loss computation of SimCLR takes $O(n^2d)$ time where $d$ is the dimensionality of the projected embeddings.

\paragraph{Non-contrastive SSL Using Asymmetric Architecture.}

Recently, researchers have started exploring non-contrastive approaches to SSL,
i.e., those that do not use negative pairs for training.
To overcome collapsed embeddings, models such as BYOL \cite{Grill2020} and SimSiam \cite{chen2021exploring}
introduce asymmetry in the architecture, e.g., by suppressing gradient updates and/or using the moving average of network parameters for one branch of the Siamese network.
These methods are heuristically motivated, as collapsed embeddings are not explicitly penalized; however, they are effective in practice.

\paragraph{Non-contrastive SSL by Decorrelating Regularization.}

A line of work exists that maintains a standard (symmetric) Siamese network but introduces loss functions to suppress collapsed embeddings.
These loss functions also have regularization terms to promote feature decorrelation.
Barlow Twins \cite{Zbontar2021} is the first method in this line.
It uses a decorrelating regularizer based on cross-correlation across two views.
VICReg \cite{Bardes2022} uses regularizers that are defined in terms of covariance matrices of individual views.
We review these methods in detail in \cref{sec:barlow-twins-vicreg}. 

\paragraph{Non-contrastive SSL by Whitening.}

Some authors \cite{ermolov2021whitening,Hua2021} have used whitening to explicitly decorrelate features during training,
as opposed to performing regularization.
Subsequent work \citet{Zhang2022} whitened both features and instances.
Because the whitening procedures used in these approaches require the computation of all the eigenvalues of the covariance matrices, a training epoch takes $O(\min(d n^2, n d^2))$ time,
which is inefficient with large $d$ or $n$.

\paragraph{Use of Convolution in Machine Learning.}

Convolution is the fundamental building block of convolution neural networks (CNNs). CNNs take (linear) convolution of input vectors with small learnable kernels to extract local features.
Although FFT reduces the asymptotic complexity of convolution computation,
it is seldom used with CNNs, because the size of kernels is typically too small to benefit from speed-up by FFT.
In contrast to CNNs, we use (circular) convolution to compute summary statistics of covariance and cross-correlation matrices.

In other areas of machine learning, circular convolution and its non-commutative analogue, circular correlation, have been used for implementing associative memory \cite{Borsellino1973,Schonemann1987,Plate2003}.
The idea has recently been applied to knowledge graph embeddings (KGEs) \cite{Nickel2016},
with the resulting model later shown \cite{Hayashi2017} to be isomorphic to complex-valued KGEs \cite{Trouillon2016} by means of the convolution theorem.

\section{SSL Using Decorrelating Regularizers}
\label{sec:barlow-twins-vicreg}

This section reviews Barlow Twins \cite{Zbontar2021}
and VICReg \cite{Bardes2022}.
Given a batch of $n$ original training examples,
these models apply two randomly chosen transformations to each example.
The transformed examples are then processed independently by a neural network to produce twin embeddings (views) of the original.
Let $ \vecA[k], \vecB[k] \in \Rset^d $ denote the twin embeddings thus obtained for the $k$th example, and let $\setA = \{ \vecA[k] \}_{k=1}^n$, $\setB = \{ \vecB[k] \}_{k=1}^n$ be the sets of embeddings for individual views.
Given these embeddings, the network is trained to minimize the loss function specific to each model.

\paragraph{Barlow Twins.}

Barlow Twins optimizes the loss function defined in terms of the cross-correlation matrix $\matCAB \in \mathbb{R}^{d \times d}$ between views $\setA$ and $\setB$:
\begin{align}
  \lossfunc_{\textup{BT}}
  & = \sum_{i=0}^{d-1} \left( 1 - [\matCAB]_{ii} \right)^2 
  + \lambda
\Rbarlow (\matCAB) ,
  \label{eq:bt-loss}
\end{align}
where hyperparameter $\lambda \ge 0$ controls the strength of regularization, and
$\Rbarlow: \Rset^{d\times d} \to \Rset$ is a regularizer function defined as
\begin{align}
  \Rbarlow (\matM) &= \sum_{i=0}^{d-1} \sum_{\substack{ j=0 \\ j \neq i}}^{d-1} [\matM]_{ij}^2 . \label{eq:bt-regularizer}
\end{align}
The first term in \cref{eq:bt-loss} is minimized when the corresponding features of two views are fully correlated, i.e., $[\matCAB]_{ii} = 1$ for $i=0,\dots,d-1$.
This term can be efficiently computed in $O(n d)$ time.
Regularizer $\Rbarlow (\matCAB)$ in the second term is responsible for feature decorrelation,
as it is minimized when all the off-diagonal elements in $\matCAB$ are zero.
For a large $d$, $\Rbarlow (\matCAB)$ can be a computational burden, as it
takes $O(n d^2)$ time to compute
due to the $d\times d$ matrix $\matCAB$.

\paragraph{VICReg.}

Let $\matKA, \matKB \in \Rset^{d \times d}$ be the covariance matrices of $\setA$ and $\setB$, respectively.
The VICReg loss function is
\begin{align}
  \lossfunc_{\textup{VIC}} 
  & = \frac{\alpha}{n} \sum_{k=1}^n \| \vecA[k] - \vecB[k] \|^2_2 \nonumber \\
  & \quad + \frac{\mu}{d} \left( \Rvar(\matKA) + \Rvar(\matKB) \right) \nonumber \\
  & \quad + \frac{\nu}{d} \left( \Rcov(\matKA) + \Rcov(\matKB) \right) ,
    \label{eq:vicreg-loss}
\end{align}
where
hyperparameters $\alpha, \mu, \nu \ge 0$ control the importance of individual terms,
and $\Rvar$ is the regularizer defined as
\begin{align}
  \Rvar (\matM) & = \sum_{i=0}^{d-1} \text{max} (0, \gamma - \sqrt{ [\matM]_{ii}} ) , \label{eq:vicreg-regularizer-variance} \end{align}
with the target standard deviation $\gamma > 0$.
Function $\Rcov$ is the same regularizer as used in Barlow Twins (\cref{eq:bt-regularizer}),
but here, it is applied to $\matKA$ and $\matKB$ instead of $\matCAB$.

The first term in \cref{eq:vicreg-loss} brings two embeddings of the same example closer.
Regularizer $\Rvar$ penalizes collapsed embeddings with zero variance, whereas $\Rcov$ promotes the diversity of features by encouraging the feature covariance to be zero.
The time complexity of calculating $\Rcov(\matKA)$ and $\Rcov(\matKB)$ is $O(n d^2)$, identical to that of $\Roff(\matCAB)$ in Barlow Twins.

\section{Proposed Method}
\label{sec:method}

We propose a weaker but efficiently computable alternative to the regularizer function $\Roff$. In the following, our regularizer is presented in terms of cross-correlation matrix $\matCAB$, similarly to Barlow Twins.
However, if applied instead to $\matKA$ and $\matKB$, it produces a relaxed version of the corresponding regularization terms in the VICReg loss.

\subsection{Regularizer Based on Sums of Cross-cor\-re\-la\-tions}

\label{sec:bt-style-regularizer}

Recall that the Barlow Twins loss is based on the cross-correlation matrix $\matC = \matCAB$ of two views $\setA = \{ \vecA[k] \}_{k=1}^n$ and $\setB = \{ \vecB[k] \}_{k=1}^n$.
For brevity, assume that both $A$ and $B$ are standardized.
Then, we can simply write $ \matC = (1/(n-1))\sum_{k=1}^n \vecA[k] { \vecB[k] }\transpose $.

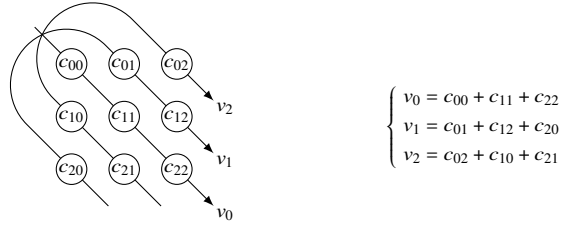
\begin{figure}[tb]
\scriptsize
  
  \def\offset{.7}
  \def\curveoffset{.35}
  \centering
  \begin{tabular}{m{.48\linewidth} @{\quad} m{.48\linewidth}}
    \begin{tikzpicture}[scale=0.7]
      \foreach \y in {0,...,2}{
\foreach \x in {0,...,2}{
          \path (\x,-\y) coordinate (n-\y-\x) ; }
        \path (3,-\y) coordinate (n-\y-3) ;
      }

\draw[-latex]
      (n-0-0) ++(-\offset,+\offset) coordinate (z0t)
      (n-2-2) ++(+\offset,-\offset) coordinate (z0b)
      (z0t) -- (z0b) ;

      \draw[latex-]
      (n-2-0)     ++(+\offset,-\offset) coordinate (z1b)
      (n-0-1)     ++(-\curveoffset,+\curveoffset) coordinate (m)
      (n-1-2)     ++(+\offset,-\offset) coordinate (z1t)
      (z1t) -- (m) arc(45:225:0.75*1.41421356) -- (z1b) ;
      
      \draw[-latex]
      (n-0-2)     ++(+\offset,-\offset) coordinate (z2t)
      (n-1-0)     ++(-\curveoffset,+\curveoffset) coordinate (m)
      (n-2-1)     ++(+\offset,-\offset) coordinate (z2b)
      (z2b) -- (m) arc(225:45:0.75*1.41421356) -- (z2t) ;

      \foreach \y in {0,...,2}{
        \foreach \x in {0,...,2}{
          \path (\x,-\y) node[draw,fill=white,circle,inner sep=.1ex] {$c_{\y\x}$} ;
        }
      }

      \node[anchor=north west, inner sep=.2ex] at (z0b.south east) {$\sumsym_0$} ;
      \node[anchor=north west, inner sep=.2ex] at (z1t.south east) {$\sumsym_1$} ;
      \node[anchor=north west, inner sep=.2ex] at (z2t.south east) {$\sumsym_2$} ;
    \end{tikzpicture}
    & \centering {
      $
      \left\{\;
      \begin{aligned}
        v_0 & = c_{00} + c_{11} + c_{22} \\
        v_1 & = c_{01} + c_{12} + c_{20} \\
        v_2 & = c_{02} + c_{10} + c_{21}
      \end{aligned}
              \right.
              $
              }
  \end{tabular}
  \caption{
    A $3\times 3$ cross-correlation matrix $\matC = \left[c_{i j}\right]$ ($i,j=0, 1, 2$) and $\sumvec(\matC) = [\sumsym_0 \; \sumsym_1 \; \sumsym_2]\transpose$.
}
  \label{fig:summary-vector}
\end{figure}

Our regularizer
is defined in terms of
a $d$-dimensional ``summary'' vector of the $d \times d$ matrix $\matC$.
This vector, denoted by $\sumvec(\matC)$, is given componentwise by
\begin{align}
  \label{eq:summary-vector}
  \left[ \sumvec (\matC) \right]_i & = \sum_{j=0}^{d-1} [\matC]_{j, (i+j) \bmod d} . \end{align}
Note the zero-based component indices. 
The zeroth component $[\sumvec(\matC)]_0$ is the trace of $\matC$. Each of the remaining $d-1$ components corresponds to a sum of $d$ different off-diagonal elements of $\matC$, with no single element appearing in two distinct sums.
Thus, every element in $\matC$ appears exactly once in the summations in \cref{eq:summary-vector}.
The calculation of a summary vector for a $3\times 3$ covariance matrix is illustrated in \cref{fig:summary-vector}.

Now, we define a regularizer in terms of all but the zeroth component of $\sumvec(\matC)$:
\begin{equation}
  \Rsum (\matC) = \sum_{i=1}^{d-1} \|[\sumvec(\matC)]_i\|_q^q ,
  \label{eq:bt-style-regularizer}
\end{equation}
where hyperparameter $q \in \{1,2\}$. This function $\Rsum$ can be used as a drop-in replacement for $\Rbarlow$ in the Barlow Twins loss (\cref{eq:bt-loss}).
The zeroth component $[\sumvec(\matC)]_0$ is excluded in \cref{eq:bt-style-regularizer},
because it concerns the diagonal elements of $\matC$ that are irrelevant to feature decorrelation;
they do not appear in Barlow Twin's regularizer $\Rbarlow(\matC)$ either.

The regularizer $\Rsum$ is weaker than $\Rbarlow$
in that it imposes constraints on the components of the summary vector, or the sums of $d$ elements of $\matC$,
whereas $\Rbarlow$ constrains individual elements.
Indeed, the minimizers of $\Rbarlow(\matC)$ also minimize $\Rsum(\matC)$, but the converse does not necessarily hold.
However,
as we discuss in \cref{sec:efficient-computation}, $\Rsum$ allows faster computation.
Furthermore, in \cref{sec:permutation}, we provide a simple technique to mitigate the weakness of $\Rsum$.

\subsection{Efficient Computation} \label{sec:efficient-computation}

Computing $\sumvec(\matC)$ by \cref{eq:summary-vector} requires the cross-correlation matrix $\matCAB$,
whose calculation incurs the same computational cost as that of the Barlow Twins loss.
However, by means of the Fourier transform,
$\sumvec(\matC)$ can be calculated directly from the vectors in $\setA$ and $\setB$ without explicit calculation of their cross-correlation matrix.
To this end, we use involution and circular convolution. 

For a vector $\mat{x} \in \Rset^d$,
its \emph{involution} \cite{Schonemann1987} (also called \emph{flipping} \citet{Smith2008})
$\flip(\mat{x})$
is the vector obtained by reversing the order of its first (not the zeroth) to $(d-1)$st components;
i.e., $[\flip(\mat{x})]_i = [\mat{x}]_{(d-i) \bmod d} $ for $i=0, \ldots, d-1$.

For vectors $\mat{x}, \mat{y} \in \Rset^d$, 
their \emph{circular convolution} $\mat{x} * \mat{y}$ is a $d$-dimensional vector with components given as
\begin{align}
  \left[ \mat{x} * \mat{y} \right]_i & = \sum_{j=0}^ {d-1} \left[ \mat{x}\, \mat{y}\transpose \right]_{j, (i-j) \bmod d} . \label{eq:circular-convolution}
\end{align}
As a result, circular convolution is known as the ``compressed outer product.''

Now,
for each twin embedding pair $\vecA[k] \in \setA$ and $\vecB[k] \in \setB$ ($k = 1, \ldots, n$),
let us consider vector $\flip(\vecA[k]) * \vecB[k] \in \Rset^d$.\footnote{
  The vector $ \flip( \mat{x} ) * \mat{y} $ is known as the \emph{circular (cross-)\linebreak[0]correlation} of $\mat{x}$ and $\mat{y}$ \cite{Schonemann1987,Plate2003,Smith2008}.
We opt not to use this term in this paper to avoid confusion with the cross-correlation of random vectors, which is used in Barlow Twins.}
Noting the indices altered by involution,
we see that this vector is given componentwise by \begin{align}
  \left[ \flip( \vecA[k] ) * \vecB[k] \right]_i
& = \sum_{j=0} ^{d-1} \left[ \vecA[k] { \vecB[k] }\transpose \right]_{j, (i+j) \bmod d} . \label{eq:circular-correlation}
\end{align}
Substituting $\matC = (1/(n-1)) \sum_{k=1}^{n} \vecA[k]{ \vecB[k] }\transpose $ into \cref{eq:summary-vector}
and using \cref{eq:circular-correlation}, we have
\begin{align} \left[ \sumvec(\matC) \right]_i 
  & = \sum_{j=0}^{d-1} { \overbrace{ \left[ \frac{1}{n-1} \sum_{k=1}^{n}  \vecA[k] { \vecB[k] }\transpose \right]}^{\matC}}_{j, (i+j) \bmod d} \nonumber\\
  & = \frac{1}{n-1} \sum_{k=1}^{n} \sum_{j=0}^{d-1} \left[  \vecA[k] { \vecB[k] }\transpose \right]_{j, (i+j) \bmod d}  \nonumber \\
  & = \frac{1}{n-1} \sum_{k=1}^{n} \left[ \flip( \vecA[k] ) * \vecB[k] \right]_i  ,
\end{align}
or, as a vector,
\begin{align}
  \sumvec(\matC) & = \frac{1}{n-1} \sum_{k=1}^{n} \flip( \vecA[k] ) * \vecB[k] . \label{eq:summary-vector-by-circular-correlation}
\end{align}
Let $\DFT$ and $\IDFT$ denote the discrete Fourier transform and inverse discrete Fourier transform, respectively.
Noting that $\DFT(\flip(\mat{x})) = \overline{\DFT(\mat{x})}$ for any $\mat{x} \in \Rset^d$ (see e.g., \citet{Smith2008}, Section~7.4.2)
and using the convolution theorem $\DFT(\mat{x} * \mat{y}) = \DFT(\mat{x}) \hprod \DFT(\mat{y})$, we have
\begin{align}
  \flip( \vecA[k] ) * \vecB[k] & = \IDFT \left(  \overline{ \DFT( \vecA[k] ) } \hprod \DFT( \vecB[k] ) \right) . \label{eq:circular-correlation-by-fft}
\end{align}
Substituting \cref{eq:circular-correlation-by-fft} into \cref{eq:summary-vector-by-circular-correlation}, we obtain
\begin{align}
  \sumvec(\matC) & = \frac{1}{n-1} \sum_{k=1}^n \overbrace { \IDFT \left(  \overline{ \DFT(\vecA[k]) } \hprod \DFT (\vecB[k]) \right) }^{ \flip( \vecA[k] ) * \vecB[k] } \nonumber \\
                 & = \frac{1}{n-1} \IDFT \left( \sum_{k=1}^n \overline{  \DFT( \vecA[k] )} \hprod \DFT(\vecB[k]) \right). \label{eq:summary-vector-by-fft}
\end{align}
Using this equation, we can compute $\sumvec(\matC)$
directly from the embedding vectors $\{ \vecA[k], \vecB[k] \}_{k=1}^n$, bypassing the cumbersome calculation of $\matC$.
Noting that the (inverse) Fourier transform of a $d$-dimensional vector can be performed in $O(d \log d)$ time by the FFT algorithm,
while the calculation of complex conjugates, component products, and the sum of $n$ vectors takes $O(n d)$ time, we can see that the overall time to obtain $\sumvec(\matC)$ is $O(n d\log d)$,
which is also the time needed to compute $\Rsum(\mat{C})$.
This is a substantial improvement over the $O(n d^2)$ computation time of $\Roff(\mat{C})$ in the Barlow Twins loss.

Furthermore, the space complexity of computing $\Rsum(\mat{C})$ is $O(n d)$,
which is optimal
if the $O(n d)$ space needed to store input vectors $\setA$ and $\setB$ is considered as part of the complexity.
In contrast, Barlow Twins requires extra $O(d^2)$ space to store $\matC$.

\subsection{Feature Permutation to Mitigate Undesirable Local Minima}
\label{sec:permutation}

As seen from \cref{eq:summary-vector}, the components of $\sumvec(\matC)$ are the sums of $d$ elements in $\matC$,
and the proposed regularizer $\Rsum$ (\cref{eq:bt-style-regularizer}) encourages these sums to be close to zero.
This is weaker than the regularizer $\Rbarlow$ (\cref{eq:bt-regularizer}) in the Barlow Twins loss, which pushes individual elements of $\matC$ towards zero.
Indeed,
$\Rsum(\matC)$ can be close to zero even if individual elements in $\matC$ are not;
the summands in \cref{eq:summary-vector} can cancel each other, since they can be either positive or negative. 
As a result, undesirable local minima develop in the parameter space, rendering our regularizer ineffective.

Here, we propose a technique to eliminate these local minima:
we randomly permute feature indices during training so that sums of different cross-correlation terms constitute $\sumvec(\matC)$.
To understand why this simple technique is effective,
consider minimizing $\Rsum(\matC)$, regarding the elements of $\matC$ as independent\footnote{
  This assumption is not unrealistic given the high capacity of modern neural networks.
}
variables.
It is easy to see that the minimum is attained by the solutions to a homogeneous system of linear equations:
\begin{align*}
  \overbrace{ \sum_{j=0}^{d-1} [\matC]_{j, (i+j) \bmod d} }^{ \left[ \sumvec (\matC) \right]_i } & = 0 , \qquad \text{for } i = 1, \ldots, d-1.
\end{align*}
This is an underdetermined system, with only $d-1$ equations but with $d(d-1)$ unknowns, namely, $[\matC]_{j\ell}$, $j,\ell=0, \ldots, d-1, j\ne \ell$.
This underdeterminacy is the cause of nontrivial solutions such that $[\matC]_{j\ell} \ne 0$,
that is, undesirable solutions in which summands with opposite signs cancel each other in an equation.

Now, by repeatedly permuting the feature indices and minimizing the loss, we effectively introduce additional equations to the system,
since permutation can produce different sets of linear equations over the unknowns,
and these new constraints eventually render nontrivial solutions inadmissible.

For ease of implementation, we permute feature indices randomly during training, instead of generating all permutations systematically at once,
similarly to the way that stochastic gradient descent is a modification to full gradient descent.
Note that the permuted feature indices need not be identical across mini-batches, even within a single epoch;
indeed, in the experiments presented in \cref{sec:experiment}, we use a different random permutation of features in every mini-batch in every epoch.

\subsection{Feature Grouping to Control the Degree of Relaxation}
\label{sec:feature-grouping}

Instead of computing a summary vector for an entire cross-correlation matrix $\mat{C}$,
we can compute summaries at a more fine-grained level. Specifically,
we partition $d$ features into groups of size $b$ each\footnote{
  If $d$ is not divisible by $b$, pad dummy features that are constantly $0$ in the last group.
}.
This partitioning induces in $\matC$ a total of $\lceil d/b \rceil^2$ block submatrices of size $b \times b$,
i.e., 
$\matC = [\matC_{i j}]$ $(i,j=1,\ldots, \lceil d/b \rceil)$ with submatrices $\matC_{i j} \in \Rset^{b \times b}$.
We then define the regularizer as
\begin{align}
  \Rsum^{(b)} (\matC) & = \sum_{i=1}^{\lceil d/b \rceil} \sum_{\ell=1}^{b-1} \left\| [\sumvec(\matC_{ii})]_\ell \right\|_q^q \nonumber \\
                      & \qquad  + \sum_{\substack{i,j=1\\i\ne j}}^{\lceil d/b \rceil} \sum_{\ell=0}^{b-1} \left \| [ \sumvec(\matC_{i j}) ]_\ell \right \|_q^q .
  \label{eq:bt-style-group-regularizer}
\end{align}

As before, $\sumvec(\matC_{i j})$ can be computed without explicitly computing $\matC_{i j}$ by means of involution, circular convolution (of subvectors of embeddings), and the Fourier transform.
Calculating a single $\sumvec(\matC_{i j})$ takes $O(n b \log b)$ time using FFT,
and since there are $\lceil d/b \rceil^2$ blocks, the total time needed to compute $\Rsum^{(b)}$ is $O((n d^2 / b) \log b )$.

The block size hyperparameter $b$ controls the granularity of the summary computation and interpolates the proposed and existing regularizers.
On one hand, the regularizer $\Rsum^{(1)}(\matC)$ reduces $\Rbarlow(\matC)$ of Barlow Twins when $b = 1$, provided that $q=2$. On the other hand, when $b=d$, we recover $\Rsum^{(d)}(\matC) = \Rsum(\matC)$ given in \cref{eq:bt-style-regularizer}.
Thus, this grouping formulation provides a generalization of Barlow Twins, with parameter $b$ controlling the trade-off between the computational efficiency and the degree of relaxed regularization.
Empirically, the performance can be slightly improved by the use of a feature group of moderate size, with no substantial degradation observed in training time and memory usage;
see \cref{sec:experiment}. 

It should be noted that the permutation and grouping of features are compatible and can be combined.

\subsection{Regularizer Based on Sums of Feature Covariances}
\label{sec:vicreg-style-regularizer-brief}

The function $\Rsum$ can also be used to define a VICReg-style regularizer based on covariance, simply by replacing $\Rcov$ with $\Rsum$ in \cref{eq:vicreg-loss},
and passing correlation matrices $\matKA$ or $\matKB$ instead of $\matC(\setA, \setB)$ as the argument.
Fast computation is possible with FFT, and
the grouping version is also straightforward;
this is described in greater detail in \appendixName.

\subsection{Summary of Proposed Models}
\label{sec:method-summary}

The loss functions of the proposed models are summarized below.
Let the function $R=\Rsum^{(b)}$ if feature grouping is used with block size $b$,
or let $R = \Rsum$ if grouping is not used.

For Barlow Twins-style cross-correlation regularization, the loss function is
\begin{align}
  \lossfunc &= \sum_i \left( 1 - [\matCAB]_{ii} \right)^2 + \lambda R (\matCAB), \label{eq:bt-style} \\
  \intertext{whereas for VICReg-style covariance regularization, we use}
  \lossfunc & = \frac{\alpha}{n} \sum_{i}\| \vecA^{(i)} - \vecB^{(i)} \|^2_2 
    + \frac{\mu}{d} \left( \Rvar(\matKA) + \Rvar(\matKB) \right) \nonumber \\
& \qquad  + \frac{\nu}{d} \left( R(\matKA) + R(\matKB) \right). \label{eq:vic-style}
\end{align}
Setting $R = \Roff$ in these formulas gives the original Barlow Twins and VICReg, when hyperparameter $q=2$.

\section{Experiments}
\label{sec:experiment}

We empirically evaluate the effect of the proposed regularizers. To be precise,
we train SSL models using the Barlow Twins--style loss function of \cref{eq:bt-style} and the VICReg-style loss function of \cref{eq:vic-style} 
and compare their performance with Barlow Twins and VICReg in downstream tasks.
The training time and memory consumption are also evaluated.
For our models, feature permutation is performed in every batch iteration except for the ablation study.

In the following, we briefly present the tasks and datasets used in the experiments.
See \crefappendix{sec:appendix-setup,sec:appendix-experiment} for the complete experimental setup, including the hyperparameter values
and the results of additional experiments.

\subsection{Tasks and Datasets}
\label{sec:setup}

Models are pretrained with images in the ImageNet dataset \cite{deng2009imagenet} or its subset, ImageNet-100~\cite{tian2020contrastive}, depending on the experiment.
ResNet-50 \cite{He2016} is used as the backbone for ImageNet, while ResNet-18 is used for ImageNet-100.

To evaluate downstream SSL performance, we follow the standard linear evaluation protocol:
After the backbone network is pretrained by a SSL method,
we train a linear classifier on top of the frozen backbone using labeled data from the ImageNet or ImageNet-100 training set. 
The resulting classifier is then evaluated by the top-1 and top-5 accuracy on the respective validation sets.

For transfer learning evaluation, we apply the pretrained models to an object detection task on
Pascal VOC07+12 \cite{everingham2010pascal}.
Following previous studies \cite{He2020,Zbontar2021,Bardes2022}, we use the trainval set of VOC2007 and VOC2012 for training and the test split of VOC2007 for testing. 
We fine-tune Faster R-CNN \cite{ren2015faster} with R50-C4.
Models are evaluated by three types of average precision (AP): AP, $\text{AP}_{50}$, and $\text{AP}_{75}$, where $\text{AP}_{x}$ signifies the intersection-over-union (IoU) threshold of $x$ \%.
We report the average scores over five trials.

\subsection{Results and Discussion}
\label{sec:result}

\begin{table}[tb]
  \caption{
    Linear evaluation accuracy (\%) on ImageNet-100 with $d=2048$.
Bold numbers indicate the best performance within each family (cross-correlation regularization, covariance regularization, or other SSL models).
    $\dag$: quoted from the solo-learn~\cite{TurrisidaCosta2022} GitHub repository as of December 28, 2022;
    $\ddag$: quoted from \citet{Zhang2022}.
}
  \label{tab:res-imagenet100}
  \centering
  \tablefont
  \begin{tabular}{l cc}
    \toprule 
    Model                                      & Top-1          & Top-5          \\
    \midrule 
    Barlow Twins$^\dag$                        & 80.16          & 95.14          \\
    Barlow Twins                               & 80.12          & \textbf{95.24} \\
    \xsum{} (Barlow Twins--style; no grouping) & 79.94          & 94.76          \\
    \xsum{} (Barlow Twins--style; $b=128$)     & \textbf{81.02} & \textbf{95.24} \\
    \midrule
    VICReg$^\dag$                              & 79.40          & \textbf{95.02} \\
    VICReg                                     & 79.30          & 94.30          \\
    \covsum{} (VICReg-style; no grouping)      & 79.20          & 94.96          \\
    \covsum{} (VICReg-style; $b=128$)          & \textbf{80.04} & 94.98          \\
    \midrule
    W-MSE$^\dag$ \cite{ermolov2021whitening}   & 69.06          & 91.22          \\ 
    Zero-FCL$^\ddag$ \cite{Zhang2022}          & 79.32          & 94.94          \\
    Zero-CL$^\ddag$ \cite{Zhang2022}           & 79.26          & 94.98          \\
NNCLR$^\dag$ \cite{dwibedi2021little}      & 80.16          & \textbf{95.30} \\
    BYOL$^\dag$ \cite{Grill2020}               & 80.32          & 94.94          \\
    MoCo V3$^\dag$ \cite{Chen2021}             & \textbf{80.36} & 94.96          \\ 
    \bottomrule
  \end{tabular}
\end{table}

\paragraph{Linear evaluation on ImageNet-100.}

\cref{tab:res-imagenet100} presents the results.
We can see that the accuracy of the proposed models is comparable to that of all existing models in the table, including Barlow Twins and VICReg, with or without feature grouping.

\paragraph{Linear evaluation on ImageNet.}

\begin{table}[tb]
  \caption{
    Linear evaluation accuracy (\%) on ImageNet;
    highest accuracy over 100 epochs of linear head training.
$d=8192$ for the proposed model, Barlow Twins, and VICReg.
$\dag$: quoted from the original papers of the respective methods;
    $\ddag$: quoted from the MoCo V3 GitHub repository.
}
  \label{tab:res-imagenet-ep1k}
  \centering
  \tablefont
  \begin{tabular}{l @{} r @{\quad} c @{\quad} c}
    \toprule 
    Model                                      & Epochs & Top-1         & Top-5         \\
    \midrule
    Barlow Twins$^\dag$                        & 1000   & \textbf{73.2} & 91.0          \\
    Barlow Twins                               & 1000   & 72.4          & 90.6          \\
    \xsum{} (Barlow Twins--style; no grouping) & 1000   & 73.0          & 91.2          \\
    \xsum{} (Barlow Twins--style; $b = 128$)   & 1000   & \textbf{73.2} & \textbf{91.3} \\
    \midrule 
    VICReg$^\dag$                              & 1000   & \textbf{73.2} & \textbf{91.1} \\
    VICReg                                     & 1000   & 72.6          & 90.9          \\
    \covsum{} (VICReg-style; no grouping)      & 1000   & 72.8          & \textbf{91.1} \\
\midrule
    W-MSE 4$^\dag$ \cite{ermolov2021whitening} & 400    & 72.6          & ---           \\
    Zero-CL$^\dag$ \cite{Zhang2022}            & 400    & 72.6          & 90.5          \\
    SimCLR$^\dag$ \cite{Chen2020}              & 1000   & 69.3          & 89.0          \\
    NNCLR$^\dag$ \cite{dwibedi2021little}      & 1000   & \textbf{75.4} & \textbf{92.3} \\
    BYOL$^\dag$ \cite{Grill2020}               & 1000   & 74.3          & 91.6          \\
    MoCo V3$^\ddag$ \cite{Chen2021}            & 1000   & 74.6          & ---           \\ 
    \bottomrule
  \end{tabular}
\end{table}

\cref{tab:res-imagenet-ep1k} presents the results. Due to the high computational cost of this large-scale experiment, we do not evaluate feature grouping with the VICReg-style regularization.
The proposed models perform slightly worse than NNCLR, BYOL, and MoCo V3, but have comparable performance to that of Barlow Twins and VICReg.

\paragraph{Transfer learning evaluation on Pascal VOC object detection.}

\begin{table}[tb]
  \caption{
    Results of transfer learning for object detection on VOC07+12.
    $\dag$: quoted from original papers;
    $\ddag$: quoted from \cite{He2020}.
  }
  \label{tab:res-transfer}
  \centering
  \tablefont
  \begin{tabular}{l @{\quad} c c c}
    \toprule 
    Model                                      
                                               & $\text{AP}_\text{50}$ & $\text{AP}$   & $\text{AP}_\text{75}$ \\
    \midrule
    Supervised$^\ddag$                         & 81.3                  & 53.5          & 58.8                  \\
    \midrule
    Barlow Twins$^\dag$                        & \textbf{82.6}         & \textbf{56.8} & \textbf{63.4}         \\
\xsum{} (Barlow Twins--style; no grouping) & 82.5                  & 55.0          & 61.1                  \\
    \midrule 
    VICReg$^\dag$                              & \textbf{82.4}         & ---           & ---                   \\
\covsum{} (VICReg-style; no grouping)      & 82.3                  & 56.1          & 62.1                  \\
    \bottomrule
  \end{tabular}
\end{table}

\cref{tab:res-transfer} presents the results of transfer learning. Again, the proposed models demonstrate performance comparable to that of Barlow Twins and VICReg.

\begin{table}[tb]
  \caption{
    Linear evaluation accuracy (\%) and the total training time on ImageNet with ResNet-50 backbone ($d=8192$). 
    The per GPU batch size is 128.
    Proposed refers to the proposed method using the Barlow Twins--style regularizer without grouping.
    Barlow Twins${}^\ast$ refers to the results quoted from \cite[Figure 2]{Zbontar2021}. }
  \label{tab:res-imagenet-time-gpu}
  \centering
  \tablefont
  \begin{tabular}{c @{\ \ } l@{\quad} c @{\quad} c @{\quad} c}
    \toprule 
    \#GPUs (Batch size)
      & Model                               & Top-1         & Top-5         & Time                            \\
    \midrule           
    8 (1024)
      & Barlow Twins$^\ast$                 & \textbf{73.2} & 91.0          & ---                             \\
      & Barlow Twins                        & 72.4          & 90.7          & \phantom{0}6d 14h \phantom{0}0m \\
      & \xsum{}                             & 73.0          & \textbf{91.3} & \phantom{0}5d 14h 58m           \\
\cmidrule(lr){1-5}
    4 (512)
      & Barlow Twins$^\ast$                 & ---           & ---           & ---                             \\
      & Barlow Twins                        & 72.1          & 90.2          & 12d 19h 30m                     \\
      & \xsum{}                             & \textbf{72.8} & \textbf{91.2} & 10d 21h \phantom{0}6m           \\
    \bottomrule
  \end{tabular}
\end{table}

\paragraph{Training time on ImageNet.}

\cref{tab:res-imagenet-time-gpu} presents the training time over 1000 epochs on ImageNet.
We tested two situations: training using 8 and 4 GPUs. 
In both situations, we set the per GPU batch size to 128, which leads to the effective batch size of 1024 for 8 GPUs and 512 for 4 GPUs.
As indicated in \cref{tab:res-imagenet-time-gpu}, the accuracy of the proposed model is comparable to that of Barlow Twins,
with a noticeable reduction in training time.\footnote{
  This experiment was conducted on a commercial cloud platform that limits a session to a maximum of three days.
  To finish training Barlow Twins with 8 GPUs for 1000 epochs, three sessions were required, whereas the proposed model required two sessions.
  The time reported in \cref{tab:res-imagenet-time-gpu} is the total run time of these sessions, which includes time (3 to 5 seconds) for reinitialization at the beginning of each session.
}
More precise evaluation of training speed is provided in the next experiment, and additional results are provided in \crefappendix{sec:resnet50,sec:time-ddp,sec:time-analysis}.

\begin{figure*}[tb]
  \centering
  \tablefont
  \begin{tabular}{l | m{.35\linewidth} | m{.35\linewidth}}
    \toprule
                               & \multicolumn1{c|}{Cross-correlation regularization}                                             & \multicolumn1{c}{Covariance regularization}                                                     \\
    \midrule                                                                                                                                                                                                     
    Elapsed time per 10 epochs & \includegraphics[scale=0.3]{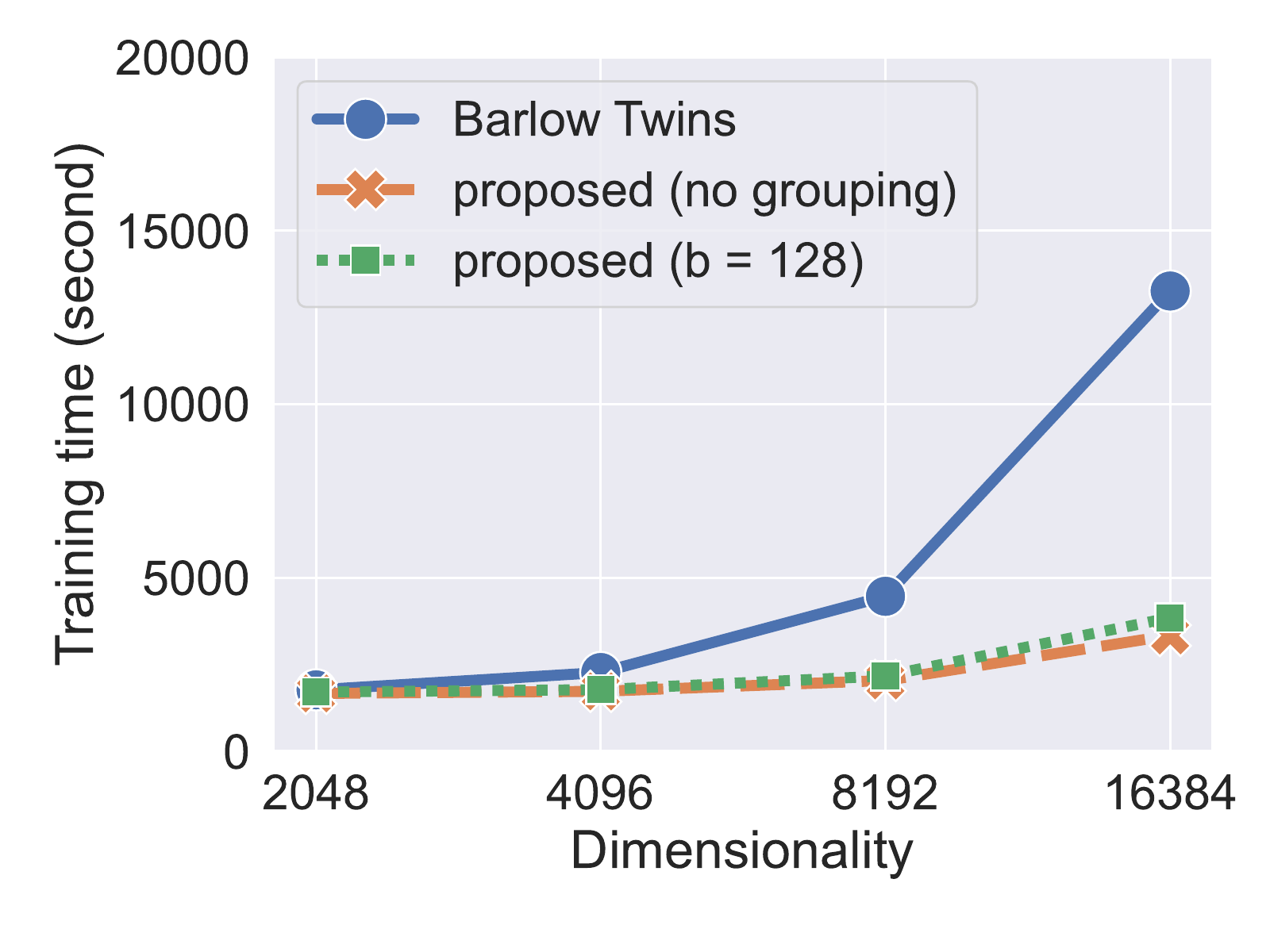} & \includegraphics[scale=0.3]{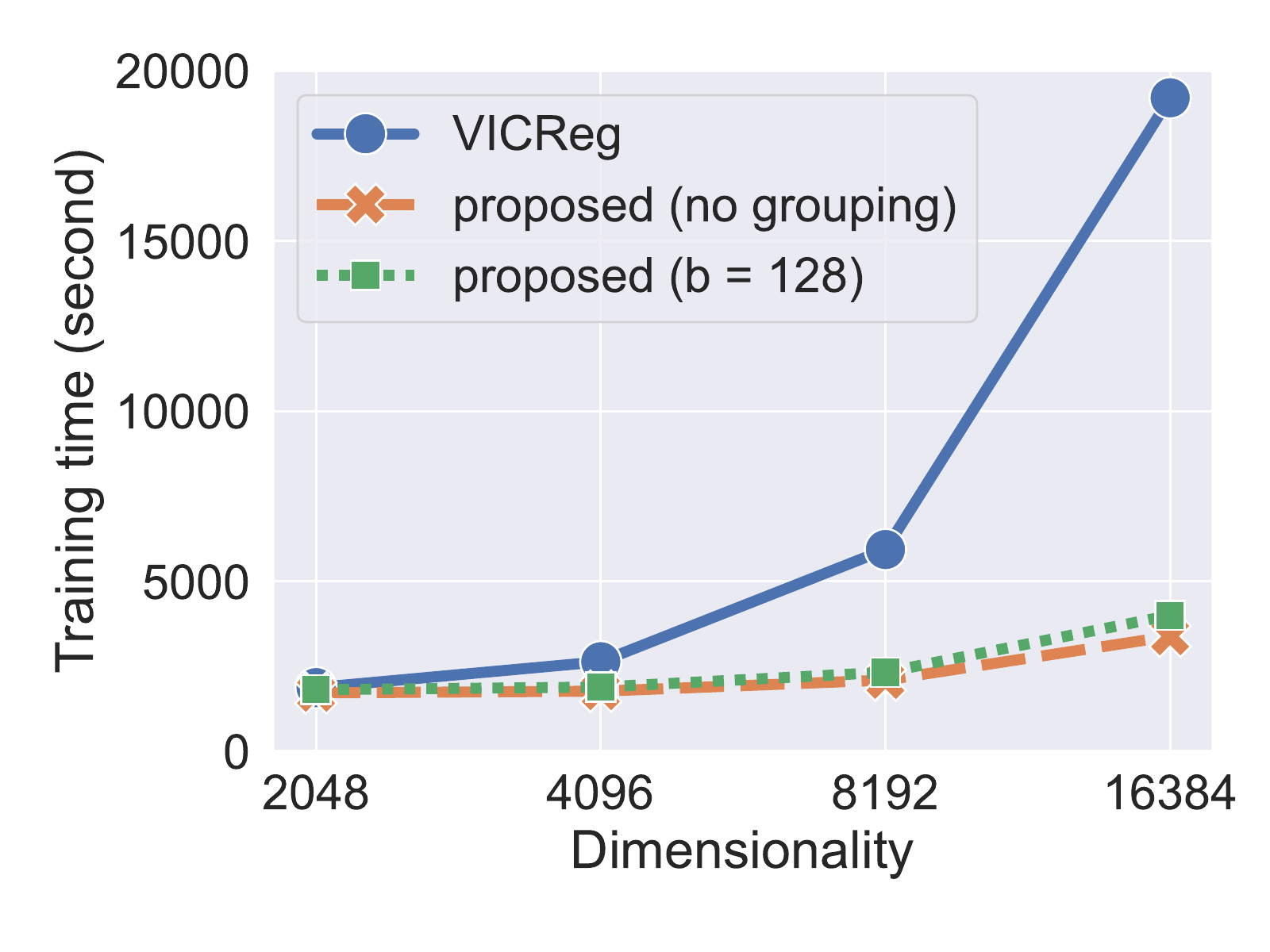} \\
    \midrule                                                                                                                                                                                                     
    Peak GPU allocated memory  & \includegraphics[scale=0.3]{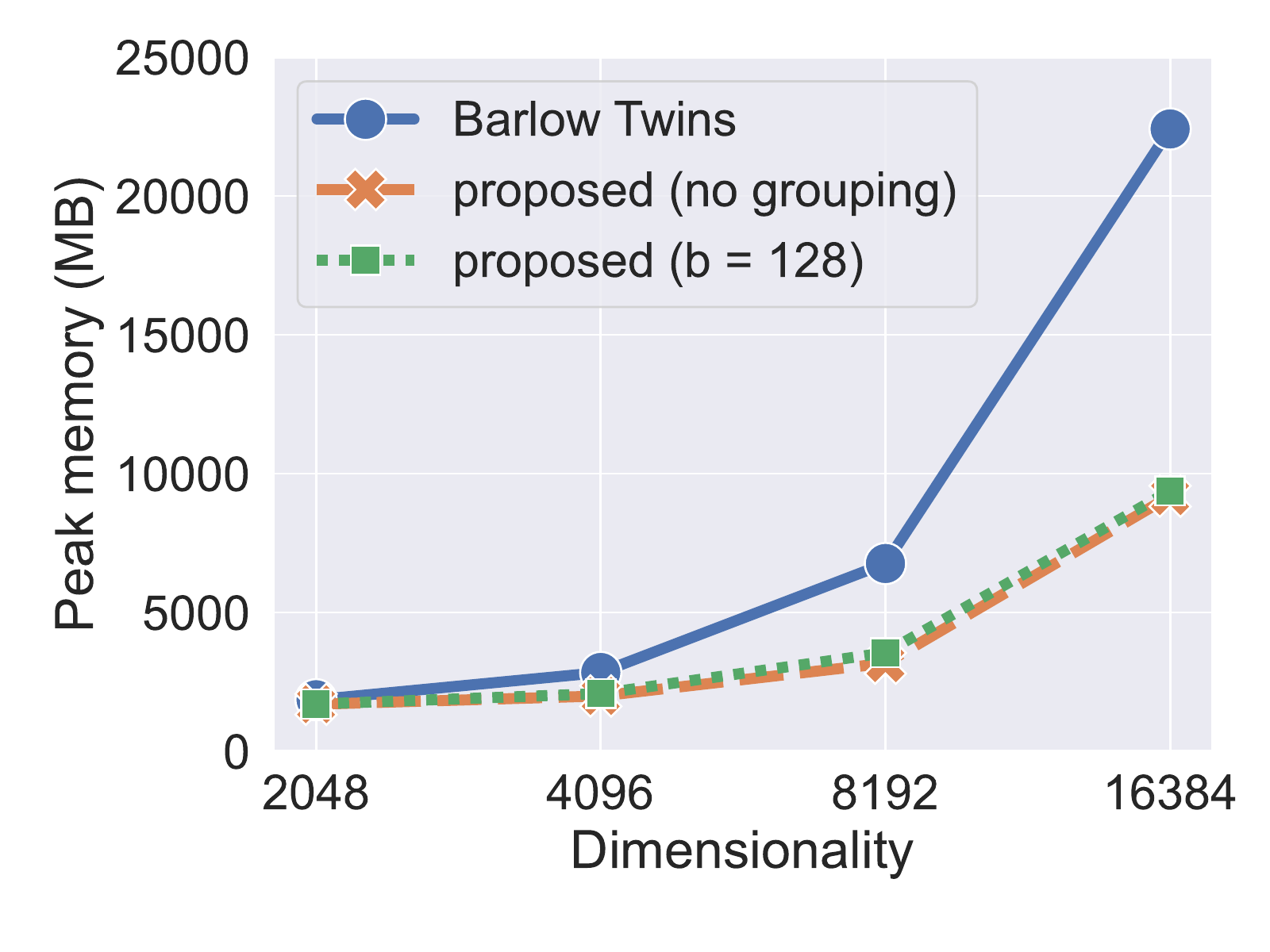}    & \includegraphics[scale=0.3]{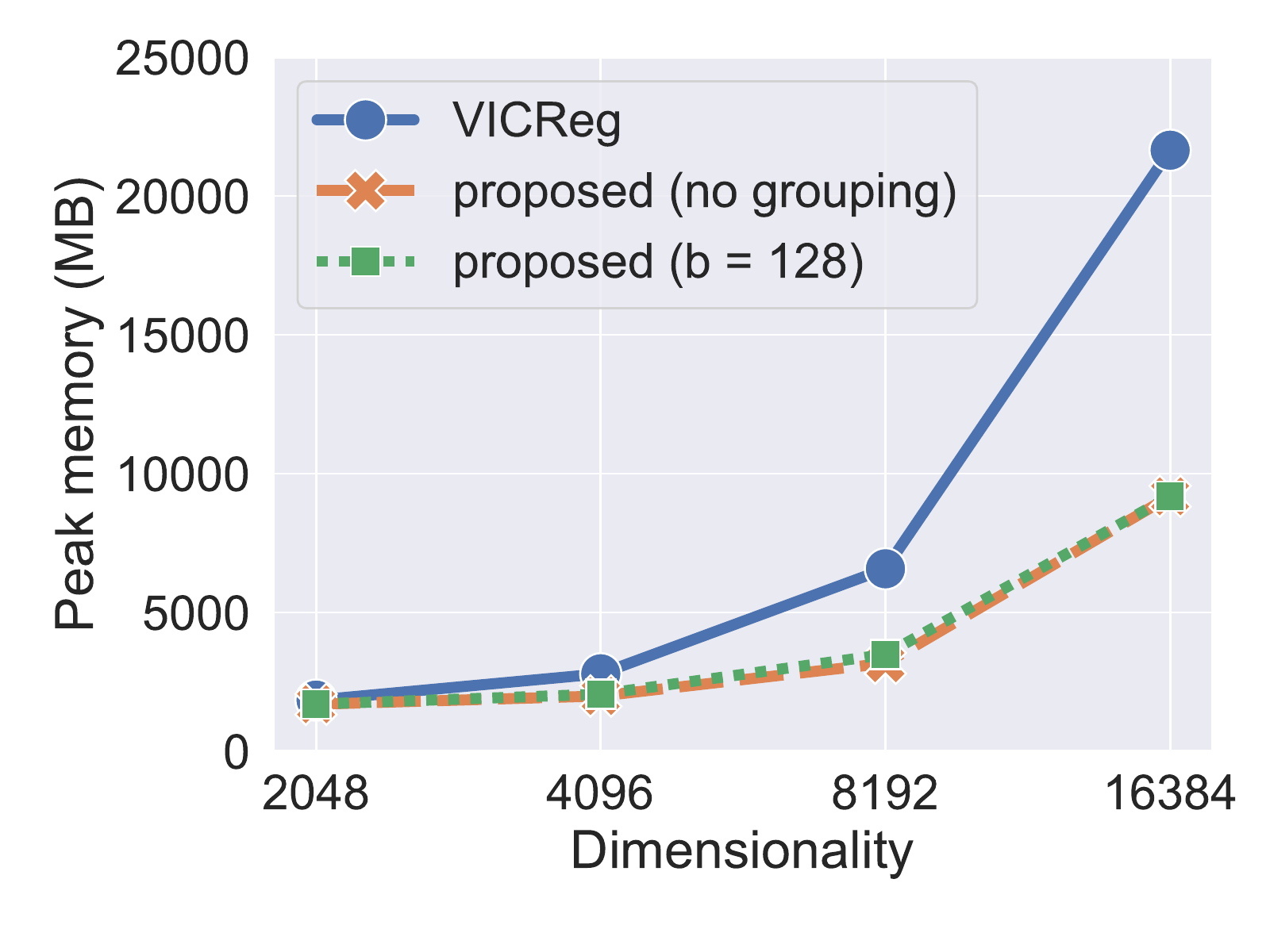}    \\
    \bottomrule
  \end{tabular}
  
  \caption{
    Training time and memory usage on ImageNet-100 with ResNet-18 on a single GPU.
}
  \label{fig:dim-time-mem-imagenet100-resnet18}
\end{figure*}

\paragraph{Dimensionality of embeddings and computational cost.}

\cref{fig:dim-time-mem-imagenet100-resnet18} presents
the elapsed time and the peak GPU memory allocation over 10 epochs on ImageNet-100 on a single GPU, with varying dimensionality of the projected embeddings: $d \in \{2048,\allowbreak 4096,\allowbreak 8192, 16384\}$. 
The improvement over Barlow Twins and VICReg becomes noticeable as $d$ is increased;
at $d=8192$, the proposed models (without grouping) are 
2.8 times as fast as VICReg, and
2.2 times as fast as Barlow Twins;
while at $d=16384$, they are 5.7 times as fast as VICReg, and
4.0 times as fast as Barlow Twins.
For both $d=8192$ and $16384$, memory consumption is reduced by more than half.
See \crefappendix{sec:resnet50,sec:time-analysis} for the results with the ResNet-50 backbone on multiple GPUs and the detail of forward/backward/loss computation time.

\begin{table}[tb]
  \caption{
    The effect of feature permutation: accuracy (\%) and training time per 10 epochs (second) on ImageNet-100.
}
  \label{tab:res-shuffle}
  \tablefont
  \centering
  \captionsetup[sub]{width=\textwidth,justification=centering}
  \subfloat[Barlow Twins--style cross-correlation regularization] {
    \tablefont
    \begin{tabular}{l c r>{\ \ }r r}
      \toprule 
\cmidrule(){3-5} 
      Grouping & Permutation & Top-1          & Top-5          & Time            \\
      \midrule
      no       & no          & 59.64          & 85.20          & \textbf{1646.2} \\
& yes         & \textbf{79.94} & \textbf{94.76} & 1668.7          \\
      \cmidrule(lr){1-5}
      $b=128$  & no          & 73.58          & 93.36          & \textbf{1697.5} \\
& yes         & \textbf{81.02} & \textbf{95.24} & 1709.6          \\
      \bottomrule
    \end{tabular}
  }
  \bigskip
  
  \subfloat[VICReg-style covariance regularization] {
    \tablefont
    \begin{tabular}{l c r r r}
      \toprule 
\cmidrule(){3-5} 
      Grouping & Permutation & Top-1          & Top-5          & Time            \\
      \midrule
      no       & no          & 57.42          & 84.26          & \textbf{1692.2} \\
& yes         & \textbf{79.20} & \textbf{94.96} & 1718.0          \\
      \cmidrule(lr){1-5}
      $b=128$  & no          & 66.26          & 89.68          & \textbf{1802.1} \\
& yes         & \textbf{80.04} & \textbf{94.98} & 1813.3          \\
      \bottomrule
    \end{tabular}
  }  
\end{table}

\paragraph{Effectiveness of feature permutation.}

\cref{tab:res-shuffle} presents the effect of feature permutation on ImageNet-100 at $d=2048$.
Regardless of whether grouping is used, the accuracy decreases significantly without permutation, 
suggesting that feature permutation is essential for the effectiveness of our proposed regularizer. 
As illustrated in the column ``Time'' in \cref{tab:res-shuffle}, the cost of permutation is negligible, even though it was performed as frequently as every batch iteration.

\begin{table}[t]
  \caption{
The regularizers of Barlow Twins and VICReg applied to the embeddings produced by the proposed models. Perm.: permutation;
    Diff: difference to the baselines.
  }
  \label{tab:res-decorrelation}
  
  \centering
  \subfloat[Barlow Twins--style cross-correlation regularization]
  {
  \tablefont
  \begin{tabular}{l @{\ } cc @{\quad} c @{\quad} c}
    \toprule
                 &          &       & Normalized Barlow                                &                \\
    Model        & Grouping & Perm. & Twins loss (\cref{eq:normalized-bt-regularizer}) & Diff           \\
    \midrule
    Barlow Twins & ---      & ---   & 0.005                                            & 0              \\
    \cmidrule(r){1-5}
    Proposed     & no       & no    & 0.564                                            & 0.559          \\
                 &          & yes   & \textbf{0.010}                                   & \textbf{0.005} \\
    \cmidrule(r){2-5}
                 & $b=128$  & no    & 0.049                                            & 0.044          \\
                 &          & yes   & \textbf{0.009}                                   & \textbf{0.004} \\
    \bottomrule
  \end{tabular}
  }
  \bigskip
  
  \subfloat[VICReg-style covariance regularization]
  {
  \tablefont
  \begin{tabular}{l @{\ } cc cc}
    \toprule 
             &          &       & Normalized                                            &           \\
    Model    & Grouping & Perm. & VICReg loss (\cref{eq:normalized-vicreg-regularizer}) & Diff      \\ 
    \midrule
    VICReg   & ---      & ---   & 0.002                                                 & 0         \\
    \cmidrule(r){1-5}
    Proposed & no       & no    & 1.999                                                 & 1.997     \\
             &          & yes   & \textbf{0.011}                                        & \bf 0.009 \\
    \cmidrule(r){2-5}
             & $b=128$  & no    & 0.379                                                 & 0.377     \\
             &          & yes   & \textbf{0.007}                                        & \bf 0.005 \\
    \bottomrule
  \end{tabular}
  }
\end{table}

We also quantitatively evaluate the degree of decorrelation obtained by feature permutation.
After the proposed models are trained,
we compute the normalized values of Barlow Twins' and VICReg's regularizers applied to the embeddings $A,B$ output by the proposed models, given as follows: \begin{align}
 & \frac{ \Rbarlow (\matCAB) }{ d(d-1) }               \label{eq:normalized-bt-regularizer} \\
 & \frac{ \Rcov(\matKA) + \Rcov(\matKB) }{ 2d(d-1) }   \label{eq:normalized-vicreg-regularizer}
\end{align}
The normalization factor $d(d-1)$ is to make the resulting values the means over $d(d-1)$ off-diagonal elements of $\matKA$, $\matKB$, and $\matCAB$.
Since VICReg has regularization terms for $\matKA$ and $\matKB$, its loss is further divided by 2. \cref{tab:res-decorrelation} presents the results on ImageNet-100 ($d$=2048).
We observe that feature permutation promotes decorrelation from the point of view of the baseline loss functions.

\begin{figure*}[tb]
  \centering
\subfloat[Top-1 accuracy]{\includegraphics[scale=0.3]{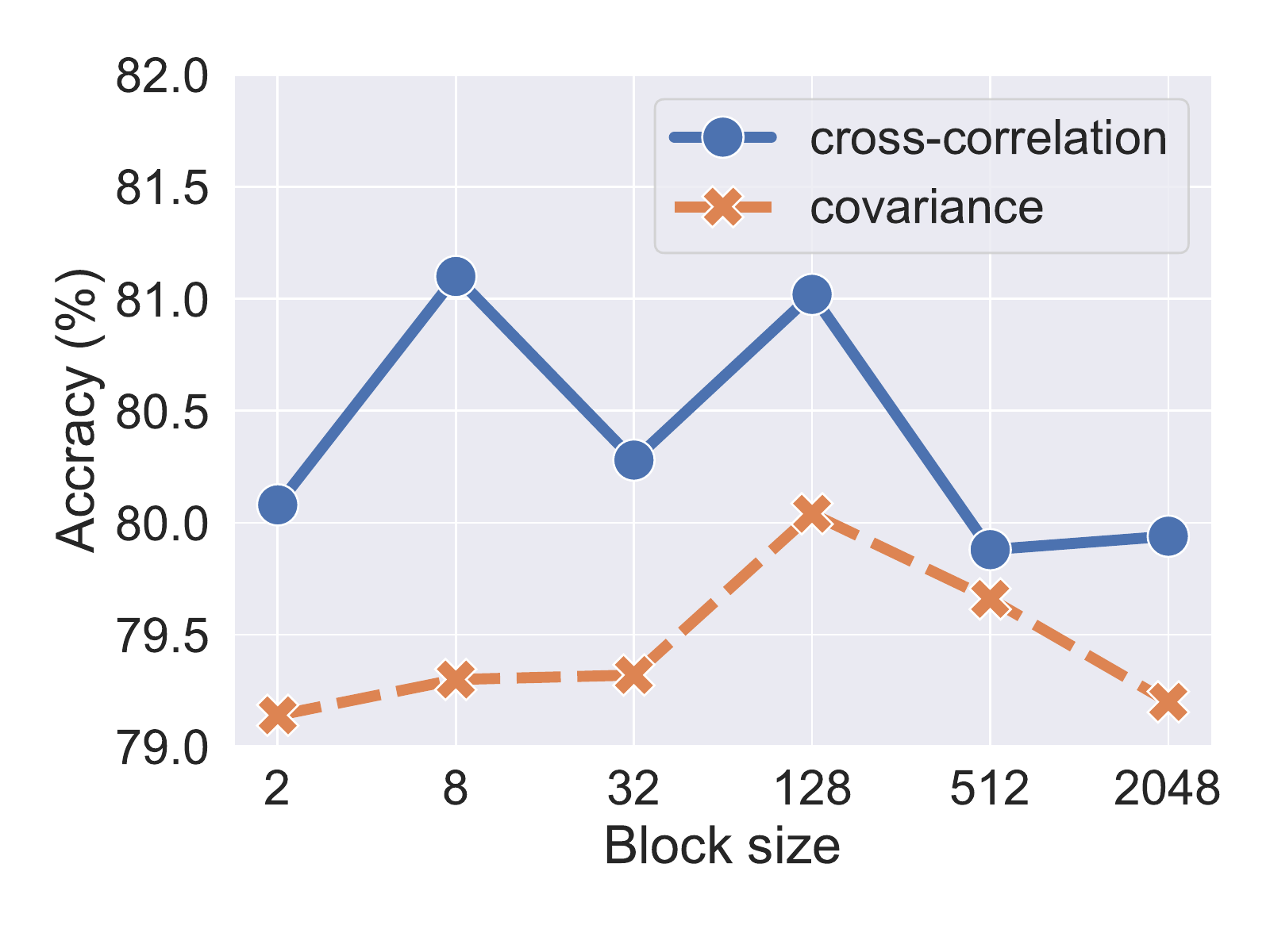}}\qquad
  \subfloat[Elapsed training time]{\includegraphics[scale=0.3]{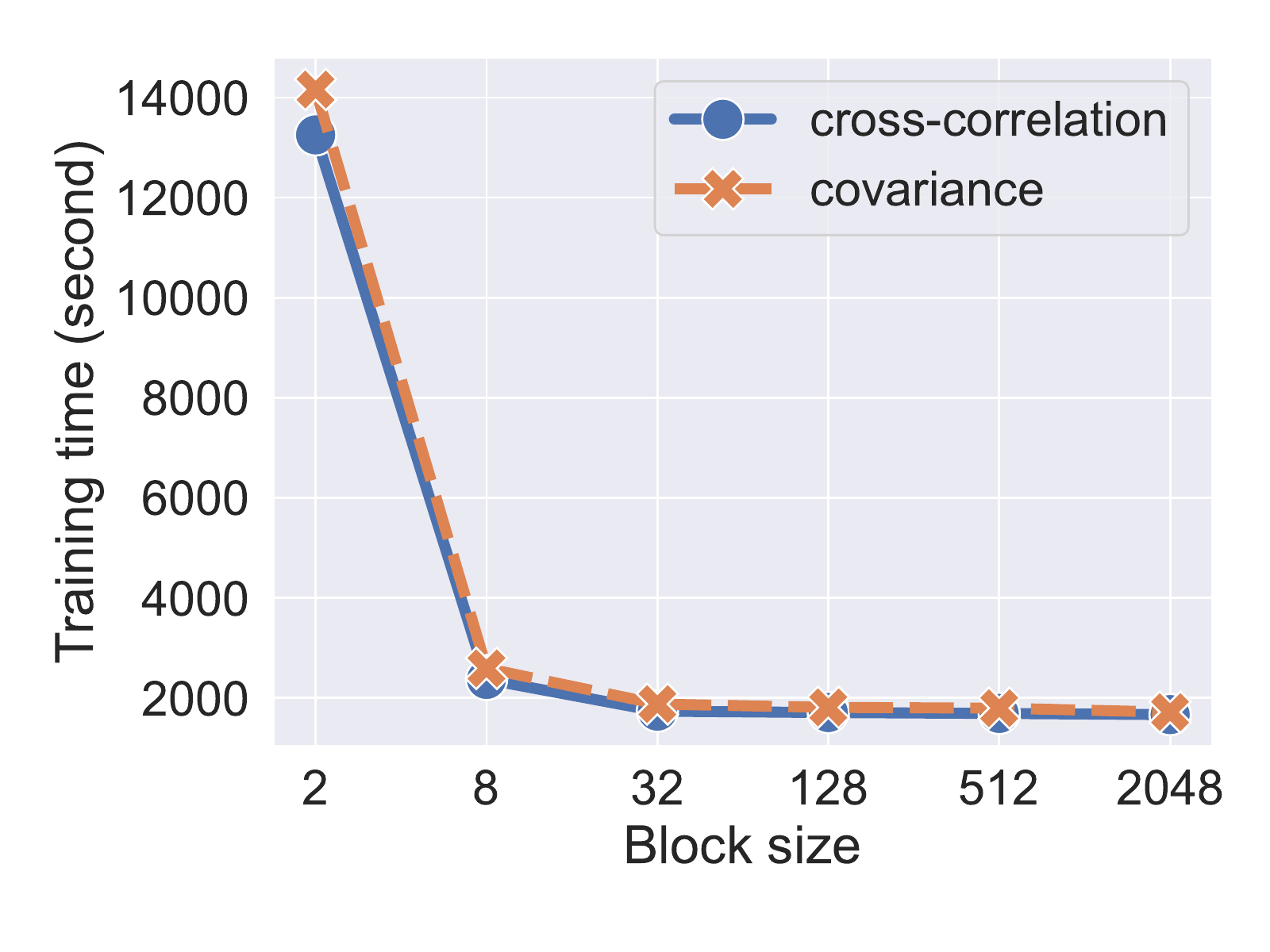}}\qquad
  \subfloat[Peak GPU memory allocated]{\includegraphics[scale=0.3]{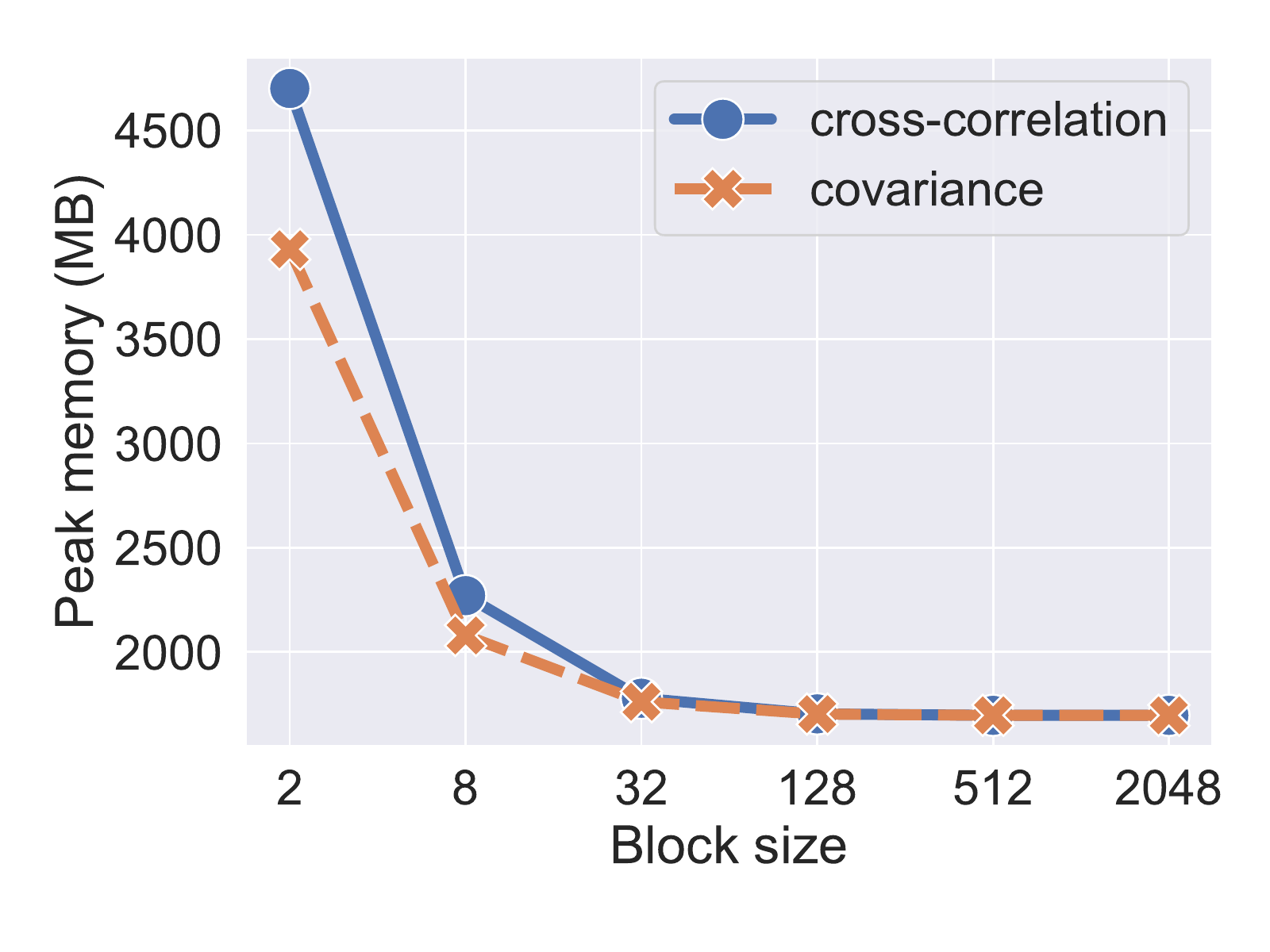}}
     
  \caption{
    The impact of the block size on ImageNet-100.
    The $x$-axis indicates the block size $b$.
}
  \label{fig:res-groupsize}
\end{figure*}

\paragraph{Impact of block size in feature grouping.}

To evaluate the effect of feature grouping on ImageNet-100,
we fix the embedding dimension at $d=2048$ and change the block size $ b \in \{2, 4, 8, ... , 2048\}$.
The block size $b = d = 2048$ corresponds to no feature grouping. 
Figure~\ref{fig:res-groupsize} presents the result,
which indicates that unless $b$ is extremely small (i.e., 8 or less), there is no significant increase in training time or GPU memory usage.
Setting $b$ to a moderate size, such as $b=128$, improves performance.

\section{Conclusion}
\label{sec:conclusion}

We have proposed a new decorrelating regularizer for non-contrastive SSL.
Given $d$-dimensional embeddings of $n$ samples, the regularizer can be computed in $O(n d\log d)$ time,
which is an improvement over the $O(n d^2)$ time required by the existing models (Barlow Twins and VICReg).
The reduced memory consumption allows for a larger batch size, which in general enables more effective models to be learned.
We also proposed a feature permutation technique to alleviate the weakness of our regularizer.

With our regularizer,
the speed-up is most notable with networks with a lightweight backbone, wherein computation at the loss node takes a large part of training time.
This makes our method an attractive option to use with knowledge-distilled backbones \cite{Hinton2015,Gao2022}.

The proposed method is not limited to SSL but potentially useful to a wider range of applications that involve decorrelation.
We plan to pursue this avenue in future work.

\iftoggle{cvprfinal}{

\section*{Acknowledgments}

We thank anonymous reviewers for their helpful comments. 
This work was partially supported by the New Energy and Industrial Technology Development Organization (NEDO) and JSPS Kakenhi Grant 19H04173.

 }{\FloatBarrier}

{\small

\bibliographystyle{ieee_fullname}
}

\FloatBarrier

\ifdefined\includeAppendix
\onecolumn
\appendix

\section{Derivation of \protect\cref{eq:circular-correlation}}
\label{sec:derivation}

Let $d$-dimensional vectors $\mat{x} = [x_0 \cdots x_{d-1}]\transpose$, $\mat{y} = [y_0 \cdots y_{d-1}]\transpose$.
We first show $ \flip( \mat{x} ) * \mat{y} = \sum_{j=0} ^{d-1} x_j y_{(i+j) \bmod d } $.
\begin{align*}
  \left[ \flip( \mat{x} ) * \mat{y} \right]_i
  & = \sum_{j=0} ^{d-1} [ \flip(\mat{x}) ]_j \, y_{(i-j) \bmod d}                                                                                        \\
  & = \sum_{j=0} ^{d-1}  x_{(d-j) \bmod d} \; y_{(i-j     ) \bmod d}          &  & \because\; [\flip(\mat{x})]_j = x_{(d-j) \bmod d} \\
  & = \sum_{j'=0}^{d-1}  x_{j' } \, y_{(i-(d-j')) \bmod d}                    &  & \because\; \text{substituting } j' = (d - j) \bmod d            \\
  & = \sum_{j'=0}^{d-1}  x_{j' } \, y_{(i+j'    ) \bmod d}                    &  & \because\; (a - d) \bmod d = a \bmod d                           \\
  & = \sum_{j=0} ^{d-1} \left[ \mat{x} { \mat{y} }\transpose \right]_{j, (i+j) \bmod d} . &  & \because\; \text{renaming variable } j' \to j                     
\end{align*}
Setting $\mat{x} = \vecA^{(k)}$ and $\mat{y} = \vecB^{(k)}$, we obtain \cref{eq:circular-correlation} in \cref{sec:efficient-computation}.
In the literature (e.g., \citeappendix{Plate2003,Smith2008}), $ \flip( \mat{x} ) * \mat{y} $ is called the \emph{circular correlation} of $\mat{x}$ and $\mat{y}$,
and the above equation is usually presented as its definition.

\section{Regularizers Based on Sums of Feature Covariances}
\label{sec:vicreg-style-regularizer}

As mentioned in \cref{sec:vicreg-style-regularizer-brief},
if we substitute $\Rsum$ for $\Rcov$ in the loss function of VICReg given in \cref{eq:vicreg-loss},
we obtain regularization based on the covariance matrices $\matKA, \matKB$ of individual views.
The resulting regularizer for $\matKA$ is:
\begin{align}
\Rsum(\matKA) & = \sum_{i=1}^{d-1} \|[\sumvec(\matKA)]_i\|_q^q , \label{eq:vicreg-style-regularizer} \end{align}
where hyperparameter $q\in \{1,2\}$.
The regularizer for $\matKB$ has the same form and is omitted.

$\Rsum(\matKA)$ can be efficiently computed by FFT, in a similar manner to $\Rsum(\matCAB)$.
For brevity, assume that set $\setA$ is centered; i.e., all features have mean $0$ in $\setA$. In this case,
its covariance matrix is $\matK(\setA) = (1/(n-1)) \sum_{k=1}^{n} \vecA^{(k)}{ \vecA^{(k)} }\transpose $.

Noting that $\DFT(\flip(\vecA^{(k)}) = \overline{\DFT(\vecA^{(k)})}$ and the convolution theorem $\DFT(\mat{x} * \mat{y}) = \DFT(\mat{x}) \circ \DFT(\mat{y})$, we have
\begin{align}
  \sumvec(\matKA) & = \frac{1}{n-1} \sum_{k=1}^n \flip( \vecA^{(k)} ) * \vecA^{(k)} \nonumber \\
                  & = \frac{1}{n-1} \sum_{k=1}^n \overbrace { \IDFT \left(  \overline{ \DFT(\vecA^{(k)}) } \hprod \DFT(\vecA^{(k)}) \right) }^{ \flip( \vecA^{(k)} ) * \vecA^{(k)} }  \nonumber \\
& = \frac{1}{n-1} \IDFT \left( \sum_{k=1}^n \overline{  \DFT(\vecA^{(k)}) } \hprod \DFT(\vecA^{(k)}) \right), \label{eq:vicreg-style-summary-vector-by-fft}
\end{align}
The grouping version is also straightforward.
Partitioning $\matKA$ into block submatrices of size $b\times b$, i.e., $\matK(A) = [\matK_{ij} ]$ ($i,j=1, \ldots, \lceil d/b \rceil$), where $\matK_{ij} \in \Rset^{b\times b}$,
and applying $\Rsum^{(b)}$ defined in \cref{eq:bt-style-group-regularizer} to it, we obtain
\begin{align}
  \Rsum^{(b)} (\matKA)   & = \sum_{i=1}^{\lceil d/b \rceil} \sum_{\ell=1}^{b-1} \| [\sumvec(\matK_{ii})]_\ell \|_q^q  + \sum_{\substack{i,j=1 \\i\ne j}}^{\lceil d/b \rceil} \sum_{\ell=0}^{b-1} \left \| [ \sumvec  (\matK_{ij} ) ]_\ell \right \|_q^q .
  \label{eq:vicreg-style-group-regularizer}
\end{align}
where block size $b$ is the hyperparameter that controls the granularity of the summary computation.
When $q=2$ and $b=d$, i.e., the block size is $(b/d)\times (b/d) = 1\times 1$, the regularizer $\Rsum^{(b)}(\matKA)$ reduces to $\Rcov(\matKA)$ of VICReg.

\section{Summary of Computational Complexity}
\label{sec:summary-complexity}

\begin{table}[tb]
  \caption{
    Complexity of loss computation. The space complexity includes $O(nd)$ memory needed to store input embeddings.
    Grouping = $b$ indicates $b$ being the size of the group (i.e., block size).
  }
  \label{tab:complexity}
  \centering
  \tablefont
  \begin{tabular}{l @{} c @{} c c}
    \toprule 
    Regularizer              & Grouping & Time                     & Space         \\
    \midrule 
    Barlow Twins             & ---      & $O(n d^2)$               & $O(nd + d^2)$ \\
    VICReg                   & ---      & $O(n d^2)$               & $O(nd + d^2)$ \\
    Proposed ($\Rsum$)       & no       & $O(n d\log d)$           & $O(n d)$      \\
    Proposed ($\Rsum^{(b)}$) & $b$      & $O((n d^2 / b) \log b )$ & $O(n d)$      \\
    \bottomrule
  \end{tabular}
\end{table}

\cref{tab:complexity} summarizes the computational complexity of the regularizers discussed in this paper.
As the table shows, the proposed regularizers are faster and cheaper than the Barlow Twins and VICReg in terms of time and space complexity.

\section{Detailed Experimental Setups}
\label{sec:appendix-setup}

All the experiments in \cref{sec:experiment} were conducted on commercial Linux servers with CUDA v11.6.2 and cuDNN v8.
We implemented our model using PyTorch v1.12.0 \citeappendix{NEURIPS2019_9015} and solo-learn v1.0.5~\citeappendix{TurrisidaCosta2022}, a library of self-supervised methods for visual representation learning built on top of PyTorch and PyTorch Lightning\footnote{\url{https://www.pytorchlightning.ai/}} v1.6.4. We also used NVIDIA DALI, a library for data loading and pre-processing to accelerate deep learning applications\footnote{\url{https://github.com/NVIDIA/DALI}}.
For object detection, detectron2 \cite{wu2019detectron2} was used.
To manage the experiments, we used Weights \& Biases, a machine learning platform for the tracking and visualization of experiments~\citeappendix{wandb2020}.
As PyTorch v1.12.0 only provides experimental support for half precision FFT\footnote{\url{https://pytorch.org/blog/pytorch-1.12-released/\#beta-complex32-and-complex-convolutions-in-pytorch}},
we trained every model with 32-bit precision, including Barlow Twins and VICReg.

\subsection{Compared Methods}

\begin{table}[tb]
  \centering
  \caption{Loss functions and regularizers in the proposed method (with and without grouping), Barlow Twins, and VICReg.
  Grouping = $b$ indicates $b$ being the size of the group (i.e., block size).}
  \label{tab:loss-functions}
  \smallskip

  {
    \captionsetup[sub]{width=\textwidth,justification=centering}

    \subfloat[Cross-correlation--regularization with Barlow Twins--style loss function][Cross-correlation--regularization with Barlow Twins--style loss function
    {$ \lossfunc = \sum_i \left( 1 - [\matCAB]_{ii} \right)^2 + \lambda R (\matCAB) $}
    \smallskip
    ]{
      \tablefont
      \begin{tabular}{l cc}
        \toprule
        Method       & Grouping & Regularizer function $R$ \\
        \midrule
        Barlow Twins & ---      & $ \Rbarlow$              \\[0.5ex]
        Proposed     & no       & $ \Rsum  $ \strut        \\[0.5ex] Proposed     & $b$      & $ \Rsum^{(b)}  $         \\
        \bottomrule
      \end{tabular}
    }
    \bigskip
    
    \subfloat[Covariance regularization with VICReg-style loss function][Covariance regularization with VICReg-style loss function
    $ \lossfunc  = \frac{\alpha}{n} \sum_{i}\| \vecA^{(i)} - \vecB^{(i)} \|^2_2
    + \frac{\mu}{d} \left( \Rvar(\matKA) + \Rvar(\matKB) \right)
    + \frac{\nu}{d} \left( R(\matKA) + R(\matKB) \right)
    $
    \smallskip
    ]{
      \tablefont
      \begin{tabular}{l cc}
        \toprule
        Method   & Grouping & Regularizer function $R$ \\
        \midrule
        VICReg   & ---      & $  \Rcov $               \\[0.5ex]
        Proposed & no       & $ \Rsum  $ \strut        \\[0.5ex]
        Proposed & $b$      & $ \Rsum^{(b)}  $         \\
        \bottomrule
      \end{tabular}
    }
  }
\end{table}

In each comparison, the proposed method and the two baselines, Barlow Twins and VICReg, used an identical network architecture, with the exception of the training loss.
The loss functions of the baselines are given by \cref{eq:bt-loss,eq:vicreg-loss},
which are repeated below as \cref{eq:bt-loss-2,eq:vicreg-loss-2} for convenience.
Let $\setA = \{\vecA^{(i)}\}_{i=1}^m$, $\setB = \{\vecB^{(i)}\}_{i=1}^m$ be the embeddings of the two views, with $i=1,\ldots, m$ indicating the original sample indices,
$\matKA, \matKB \in \Rset^{d\times d}$ be their respective covariance matrices, and $\matCAB \in \Rset^{d\times d}$ be the cross-correlation matrices between $A$ and $B$.
\begin{align}
\lossfunc_{\textup{BT}}
  & = \sum_{i=0}^{d-1} \left( 1 - [\matCAB]_{ii} \right)^2 
  + \lambda
\Rbarlow (\matCAB) , \label{eq:bt-loss-2} \\
  \lossfunc_{\textup{VIC}} 
  & = \frac{\alpha}{n} \sum_{i = 1}^m \| \vecA^{(i)} - \vecB^{(i)} \|^2_2 + \frac{\mu}{d} \left( \Rvar(\matKA) + \Rvar(\matKB) \right) + \frac{\nu}{d} \left( \Rcov(\matKA) + \Rcov(\matKB) \right) , \label{eq:vicreg-loss-2}
\end{align}
where hyperparameters $\alpha, \mu, \nu, \lambda \ge 0$ determine the importance of individual terms, and the regularization functions are given by
\begin{align*}
  \Roff (\matM) & = \sum_{i=0}^{d-1} \sum_{\substack{ j=0 \\ j \neq i}}^{d-1} [\matM]_{ij}^2, \\
  \Rvar (\matM) & = \sum_{i=0}^{d-1} \text{max} (0, \gamma - \sqrt{ [\matM]_{ii}} ) . \end{align*}
For the proposed method, we replace all occurrences of $\Roff$ in
\crefappendix{eq:bt-loss-2,eq:vicreg-loss-2} with either $\Rsum$ 
(\cref{eq:bt-style-regularizer}) or $\Rsum^{(b)}$ (\cref{eq:bt-style-group-regularizer}) depending on whether grouping is used.
These functions are repeated below.
\begin{align}
\Rsum(\matM)        & = \sum_{i=1}^{d-1} \|[\sumvec(\matM)]_i\|_q^q , \label{eq:regularizer-covariance-2}                   \\
  \Rsum^{(b)} (\matM) & = \sum_{i=1}^{\lceil d/b \rceil} \sum_{\ell=1}^{b-1} \| [\sumvec(\matM_{ii})]_\ell \|_q^q
                        + \sum_{\substack{i,j=1\\i\ne j}}^{\lceil d/b \rceil} \sum_{\ell=0}^{b-1} \| [ \sumvec(\matM_{ij}) ]_\ell \|_q^q , \label{eq:bt-style-group-regularizer-2}
\end{align}
where $\matM = [\matM_{ij}]$ is a block matrix with block elements $\matM_{ij} \in \Rset^{b\times b}$, $i,j=1, \ldots, \lceil d/b \rceil$.

\cref{tab:loss-functions} summarizes the regularizers and loss functions for Barlow Twins, VICReg, and the proposed models.

\subsection{Data Augmentation}

Following the Barlow Twins paper \citeappendix{Zbontar2021}, we used non-symmetric parameters for Barlow Twins-style objectives.
For VICReg-style objectives in ImageNet-100 experiments, we used the symmetrized augmentation pipeline reported in the VICReg paper \cite[Appendix C.1]{Bardes2022}.
Following a comment in the VICReg GitHub repository\footnote{\url{https://github.com/facebookresearch/vicreg/issues/3}}, we used the non-symmetric augmentation parameters (the same parameters as in Barlow Twins) for ImageNet experiments.

\subsection{Hyperparameters}

\paragraph{ImageNet-100.}
For ImageNet-100 experiments, we followed the optimization procedure described in \citeappendix{TurrisidaCosta2022}.
To train models, stochastic gradient descent (SGD) was used with the LARS optimizer \citeappendix{you2017large} for 400 epochs. 
We used linear warm-up with cosine annealing decay for the learning rate scheduler. 
The batch size was wet to 256 (per GPU batch size is 32).
The loss scaling value and the importance coefficients for the proposed regularizers ($\nu$ in \cref{eq:vicreg-loss} and $\lambda$ in \cref{eq:bt-loss}) were sought by grid search.
In addition to these parameters, we further tuned the block size and $q$ in our regularizers (\cref{eq:bt-style-regularizer,eq:bt-style-group-regularizer}).

In linear evaluation, linear classifiers was optimized with SGD for 100 epochs. 
In training for ImageNet-100, we used the hyperparameters provided by the solo-learn library.

\cref{tab:hyperparameter} summarizes the hyperparameters for ImageNet-100 experiments.

\begin{table*}[tb]
  \centering
  \tablefont
  \caption{The hyperparameters for ImageNet-100 experiments that were used for training models. 
    The hyperparameters for Barlow Twins and VICReg were set to the values reported by \protect\citetappendix{TurrisidaCosta2022}.
    For the proposed methods, we found values by grid search.  
  }
  \label{tab:hyperparameter}
  \smallskip

  {
    \captionsetup[subfloat]{width=\textwidth,justification=centering}

    \subfloat[Cross-correlation regularization][Cross-correlation regularization with Barlow Twins--style loss function
    \smallskip
    ]{
      \tablefont
      \begin{tabular}{l cccc}
        \toprule
        Method                        & Grouping & Loss scale & $q$ & $\lambda$ \\
        \midrule
        Barlow Twins                  & ---      & 0.1        & --- & 0.0051    \\[0.5ex]
        Proposed (Barlow Twins--style) & no       & 0.125      & 2   & $2^{-10}$ \\[0.5ex] Proposed (Barlow Twins--style) & $b=128$  & 0.125      & 2   & $2^{-10}$ \\
        \bottomrule
      \end{tabular}
    }
    \bigskip
    
    \subfloat[Covariance regularization][Covariance regularization
    \smallskip
    ]{
      \tablefont
      \begin{tabular}{l cccc}
        \toprule
        Method                 & Grouping & Loss scale & $q$ & $\nu$ \\
        \midrule
        VICReg                 & ---      & ---        & --- & 1.0   \\[0.5ex]
        Proposed (VICReg-style) & no       & 0.25       & 1   & 1.0   \\[0.5ex]
        Proposed (VICReg-style) & $b=128$  & 0.25       & 1   & 2.0   \\
        \bottomrule
      \end{tabular}
    }
    \bigskip
    
    \subfloat[All other hyperparameters were set to the values reported][All other hyperparameters were set to the values reported by \protect\citetappendix{TurrisidaCosta2022}.
    \smallskip
    ]{
      \tablefont
      \begin{tabular}{ccccccc}
        \toprule
Learning rate  & Weight decay & Batch size & Warmup epochs & $\alpha$ & $\mu$                               \\
        \midrule
0.3           & $ 10^{-4} $  & 256        & 10             & 25.0     & 25.0                                \\
        \bottomrule
      \end{tabular}
    }
    \bigskip
    
    \subfloat[Hyperparameters for linear evaluation on ImageNet-100][Hyperparameters for linear evaluation on ImageNet-100.
    \smallskip
    ]{
      \tablefont
      \begin{tabular}{lcccc}
        \toprule
        Pretrained model                             & Learning rate & Steps for learning rate decay & Weight decay & Batch size \\
        \midrule
        Barlow Twins / Proposed (Barlow Twins--style) & 0.1           & [60, 80]                      & 0            & 256        \\
        VICReg / Proposed (VICReg-style)              & 0.3           & [60, 80]                      & 0            & 512        \\
        \bottomrule
      \end{tabular}
    }

}
\end{table*}

\paragraph{ImageNet.}
For ImageNet experiments, we followed the optimization procedure described in \citet{Zbontar2021,Bardes2022}.
We used SGD with the LARS optimizer for 1000 epochs and linear warmup with cosine annealing decay. 
We set the batch size to 1024 (per GPU batch size is 128) and used a learning rate of 0.25 by reference to the Barlow Twins GitHub repository\footnote{\url{https://github.com/facebookresearch/barlowtwins/issues/7}}. 
We searched for $\lambda$ and $q$ for the proposed regularizers by grid search.

In the linear evaluation on ImageNet, the linear head was trained for 100 epochs with SGD and cosine learning rate decay.
We tuned the learning rate and batch size for the proposed methods.
For Barlow Twins and VICReg, the learning rate and batch size are set to the values reported in the original papers \citet{Zbontar2021,Bardes2022}.

In object detection, we trained a Faster R-CNN with a C-4 backbone for 24K iterations.
The backbone is initialized with the pretrained ResNet-50 backbone.
Following \cite{He2020,Zbontar2021,Bardes2022}, we set the batch size to 16 (per GPU batch size is 2) and used a step learning rate decay (divided by 10 at 18K and 22K iterations) with a linear warmup (slope of 0.333 for 1K iterations). 
We tuned the learning rate and the region proposal network loss weight for the proposed methods.

\cref{tab:hyperparameter-imagenet} summarizes the hyperparameters for ImageNet experiments.

\begin{table*}[tb]
  \centering
  \tablefont
  \caption{The hyperparameters for ImageNet experiments that were used for training models. 
}
  \label{tab:hyperparameter-imagenet}
  \smallskip
  {
    \captionsetup[subfloat]{width=\linewidth,justification=centering}
    \subfloat[Cross-correlation regularization][Cross-correlation regularization\smallskip
    ]{
      \tablefont
      \begin{tabular}{l cccc}
        \toprule
        Method                        & Grouping & Loss scale & $q$ & $\lambda$ \\
        \midrule
        Barlow Twins                  & ---      & 0.024      & --- & 0.0051    \\[0.5ex]
        Proposed (Barlow Twins--style) & no       & 0.024      & 2   & $2^{-11}$ \\[0.5ex] Proposed (Barlow Twins--style) & $b=128$  & 0.024      & 2   & $2^{-11}$ \\[0.5ex] \bottomrule
      \end{tabular}
    }
    \bigskip
    
    \subfloat[Covariance regularization][Covariance regularization\smallskip
    ]{
      \tablefont
      \begin{tabular}{l cccccc}
        \toprule
        Method                 & Grouping & Loss scale & $q$ & $\alpha$ & $\mu$ & $\nu$ \\
        \midrule
        VICReg                 & ---      & ---        & --- & 25.0     & 25.0  & 1.0   \\[0.5ex]
        Proposed (VICReg-style) & no       & ---        & 1   & 2.5      & 2.5   & 0.1   \\[0.5ex]
        \bottomrule
      \end{tabular}
    }
    \bigskip

    \subfloat[Hyperparameters for optimization][Hyperparameters for optimization.
    \smallskip
    ]{
      \tablefont
      \begin{tabular}{ccccccc}
        \toprule
Learning rate  & Weight decay & Batch size & Warmup epochs                                                 \\
        \midrule
         0.25          & $ 10^{-6} $  & 1024       & 10                                                            \\
        \bottomrule
      \end{tabular}
    }
    \bigskip
    
    \subfloat[Hyperparameters for linear evaluation on ImageNet][Hyperparameters for linear evaluation on ImageNet.
    \smallskip
    ]{
      \tablefont
      \begin{tabular}{lcccc}
        \toprule
        Pretrained model              & Learning rate & Learning rate decay & Weight decay & Batch size \\
        \midrule
        Barlow Twins                  & 0.3           & cosine decay        & $ 10^{-6} $  & 256        \\
        VICReg                        & 0.02          & cosine decay        & $ 10^{-6} $  & 256        \\
        Proposed (Barlow Twins--style) & 0.125         & cosine decay        & $ 10^{-6} $  & 2048       \\
        Proposed (VICReg-style)        & 0.125         & cosine decay        & $ 10^{-6} $  & 256        \\
        \bottomrule
      \end{tabular}
    }
    \bigskip
    
    \subfloat[Hyperparameters for object detection on VOC07+12][Hyperparameters for object detection on VOC07+12.
    \smallskip
    ]{
      \tablefont
      \begin{tabular}{lcc}
        \toprule
        Pretrained model             & Learning rate & Region proposal network loss weigh \\
        \midrule
        \xsum{} (Barlow Twins--style) & 0.125         & 0.03125                            \\
        \covsum{} (VICReg-style)      & 0.125         & 0.125                              \\
        \bottomrule
      \end{tabular}
    }
  }
\end{table*}

\subsection{Evaluation of Training Time and Memory Consumption}
\label{sec:time-and-memory-measurement}

To discuss empirical complexity, we measured the elapsed time and peak GPU memory allocation over ten epochs.
We conducted three trials and reported the average time and memory allocation.
To avoid communication overhead between GPUs, we evaluated the results of single GPU training (not distributed data parallel training) unless otherwise noted.
The batch size was set to 32 for ImageNet-100 and 128 for ImageNet as in pretraining settings (per GPU batch size is 32 and 128).
The number of workers (the argument ``num\_workers'' in the solo-learn library used for implementation) is set to 32 for ImageNet-100 and 4 for ImageNet.

We used the Simple Profiler in PyTorch Lightning\footnote{\url{https://pytorch-lightning.readthedocs.io/en/1.6.4/advanced/profiler.html\#simple-profiler}} to measure training time. 
From the profiler output, the value of the ``Total time (s)'' in the ``run\_training\_epoch'' line was extracted and plotted as the training time in \cref{fig:res-groupsize,fig:dim-time-imagenet-resnet50-ddp-multinode,fig:dim-time-mem-imagenet100-resnet18,fig:dim-time-mem-imagenet-resnet50,fig:dim-time-imagenet100-resnet18-ddp,fig:dim-time-imagenet-resnet50-ddp,fig:dim-time-imagenet-resnet50-ddp}.
We used a function in PyTorch\footnote{\url{https://pytorch.org/docs/stable/generated/torch.cuda.memory_summary.html}} to monitor memory occupied by tensors.
The value of ``Peak Usage'' in the ``Allocated memory'' line was used as the peak GPU memory allocation.

As regards the total training time in \cref{tab:res-imagenet-time-gpu}, we reported the runtime counted by WandB.\footnote{We used the value of the ``Runtime'' in WandB.}  
As mentioned in the footnote of \cref{sec:result}, our experiment was performed on a commercial cloud platform that terminates a session after three days.
To finish training Barlow Twins and VICReg with 8 GPUs for 1000 epochs on ImageNet, we needed three sessions, and the proposed model only two (see \cref{tab:res-imagenet-time-gpu}).
The timing reported in \cref{tab:res-imagenet-time-gpu} is the run time of these sessions that includes the time for initialization at the beginning of each session.
This initialization takes only 3--5 seconds at each session and does not affect the trend observed in the table.
Note that in addition to this initialization, data copy takes about 15 minutes at the start of a session, but this has already been excluded from the timing in \cref{tab:res-imagenet-time-gpu}.

\subsection{Computational Resources}

We used a cloud computing platform for the experiments. 
In the main experiments, we trained models with eight Nvidia A100-SXM4 GPUs on this platform.  
We used a single Nvidia A100 GPU to evaluate empirical complexity, except where noted.

\subsection{License of the Assets}

PyTorch has a BSD-style license\footnote{\url{https://github.com/pytorch/pytorch/blob/master/LICENSE}}. 
Solo-learn has an MIT license\footnote{\url{https://github.com/vturrisi/solo-learn/blob/main/LICENSE}}.
PyTorch Lightning is licensed under the Apache License 2.0\footnote{\url{https://github.com/Lightning-AI/lightning/blob/master/LICENSE}}.
The ImageNet\footnote{\url{https://www.image-net.org/}} dataset is publicly available and frequently used as the benchmark dataset.
The category list of ImageNet-100 is also publicly available\footnote{\url{https://github.com/HobbitLong/CMC/blob/master/imagenet100.txt}}.

\section{Additional Experiments}
\label{sec:appendix-experiment}

\subsection{\texorpdfstring{Impact of Hyperparameter $q$}{Impact of Hyperparameter q}}
\label{sec:lp-loss}

\begin{table*}[t]
  \caption{
    The accuracy with $q \in \{1, 2\}$ on ImageNet-100. 
  }
  \label{tab:res-lp}
  \centering
  \tablefont
  \begin{tabular}{lcc cc}
    \toprule 
    Model                        & Grouping & $q$ & Top-1          & Top-5          \\
    \midrule
    \xsum{} (Barlow Twins--style) & no       & 1   & 75.94          & 94.28          \\
                                 &          & 2   & \textbf{79.94} & \textbf{94.76} \\
    \cmidrule{2-5}
                                 & $b=128$  & 1   & 76.44          & 94.46          \\
                                 &          & 2   & \textbf{81.02} & \textbf{95.24} \\
    \midrule
    \covsum{} (VICReg-style)      & no       & 1   & \textbf{79.20} & \textbf{94.96} \\
                                 &          & 2   & 57.98          & 84.56          \\
    \cmidrule{2-5}
                                 & $b=128$  & 1   & \textbf{80.04} & \textbf{94.98} \\   
                                 &          & 2   & 71.78          & 92.54          \\
    \bottomrule
  \end{tabular}
\end{table*}

We investigate the effect of the hyperparameter $q$ in our regularizers (\cref{eq:bt-style-regularizer,eq:bt-style-group-regularizer}). 
\cref{tab:res-lp} shows the results with $q \in \{1, 2\}$ on ImageNet-100 ($d=2048$).
The results indicate that $q=1$ works better than $q=2$ in VICReg-style covariance regularizers. 
Conversely, $q=2$ works well in Barlow Twins--style cross-correlation regularizers.

\subsection{Training Time with ResNet-50 Backbone}
\label{sec:resnet50}

\begin{figure}[tb]
  \centering
  \tablefont
  \begin{tabular}{l | m{.35\linewidth} | m{.35\linewidth}}
    \toprule
             & \multicolumn1{c|}{Cross-correlation regularization}                                   & \multicolumn1{c}{Covariance regularization}                                          \\
    \midrule                                                                                                
    Elapsed time per 10 epochs  & \includegraphics[scale=0.3]{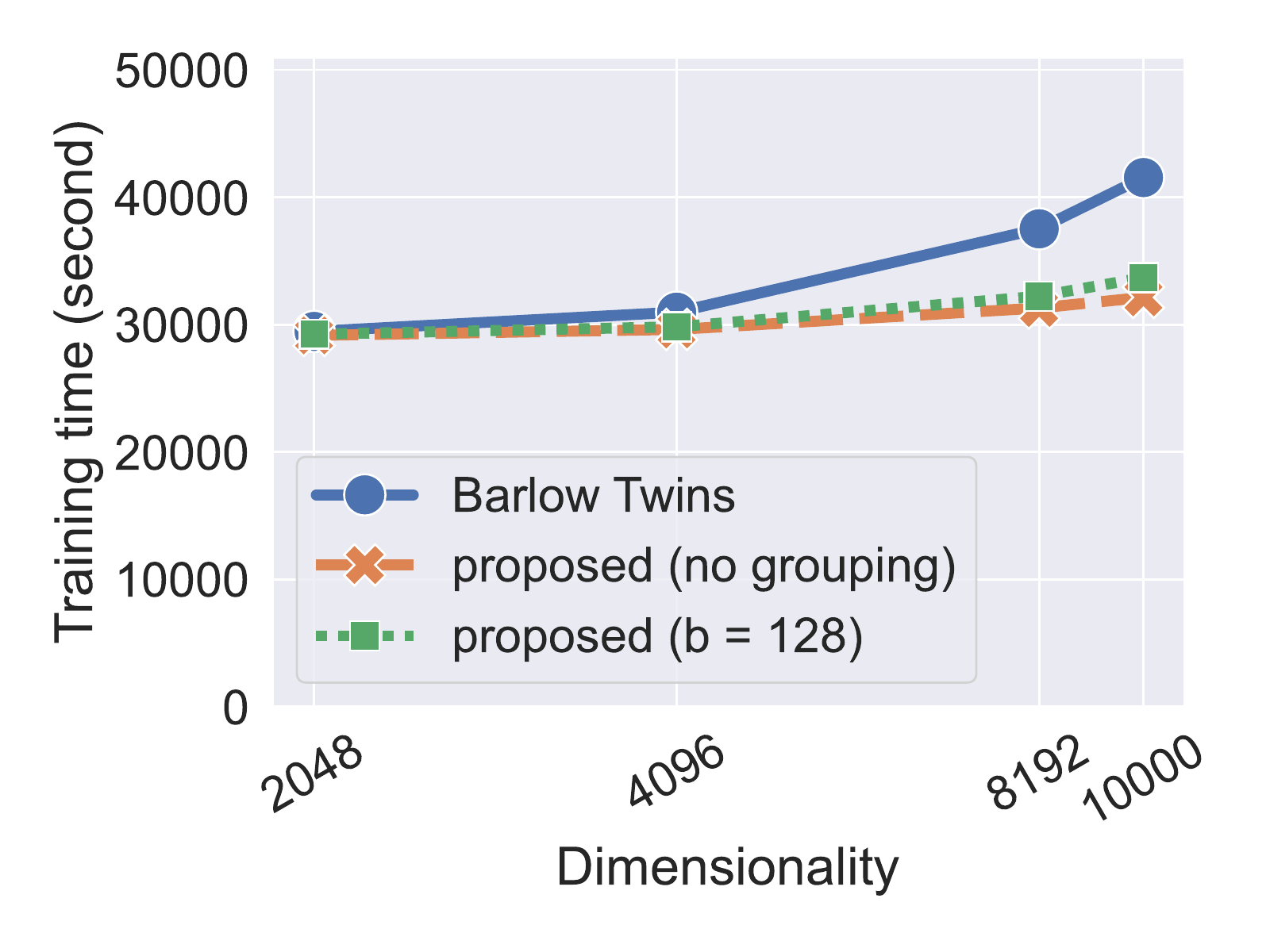} & \includegraphics[scale=0.3]{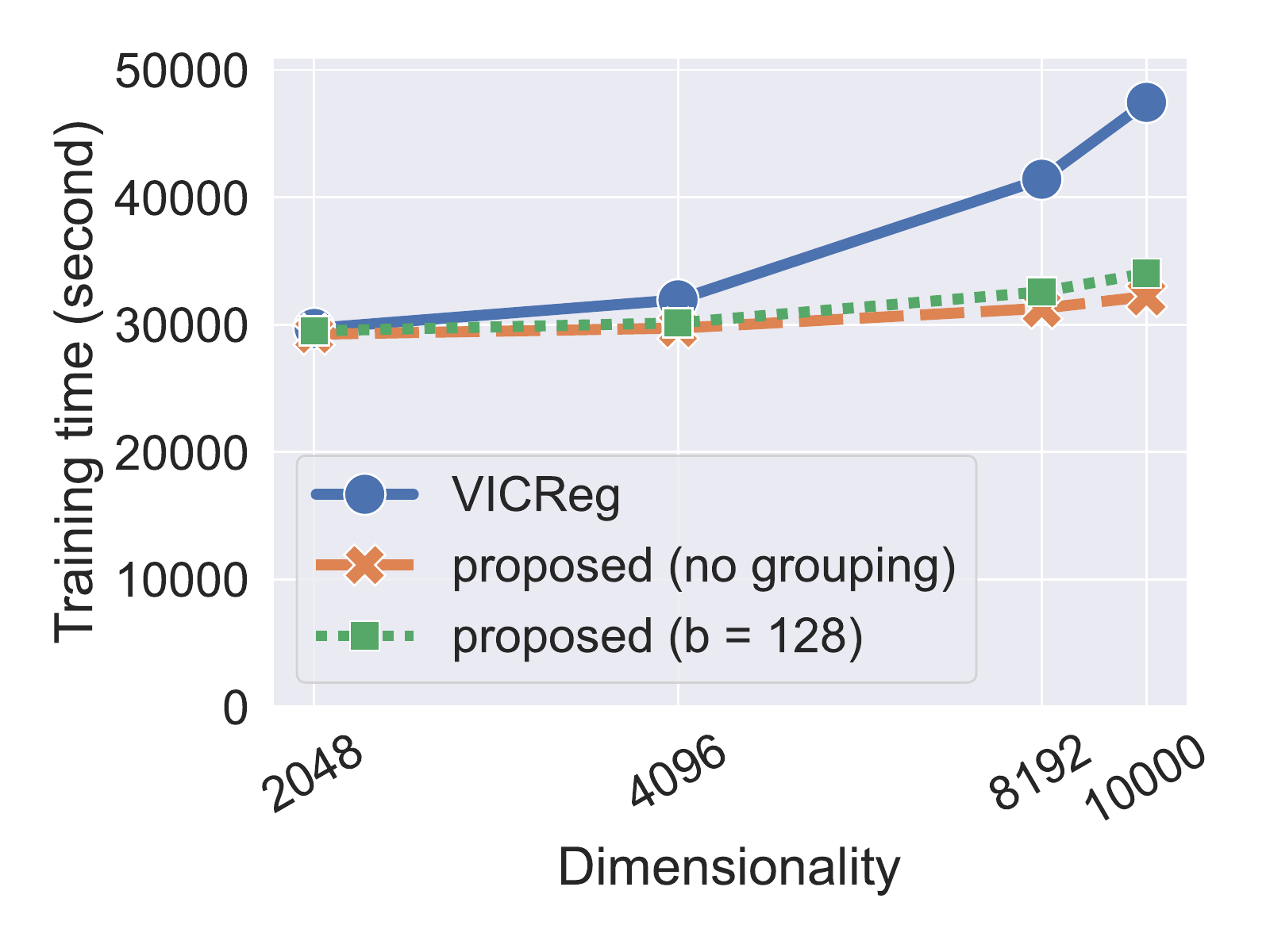} \\[-1ex]
    \midrule                                                                                                
    Peak GPU allocated memory   & \includegraphics[scale=0.3]{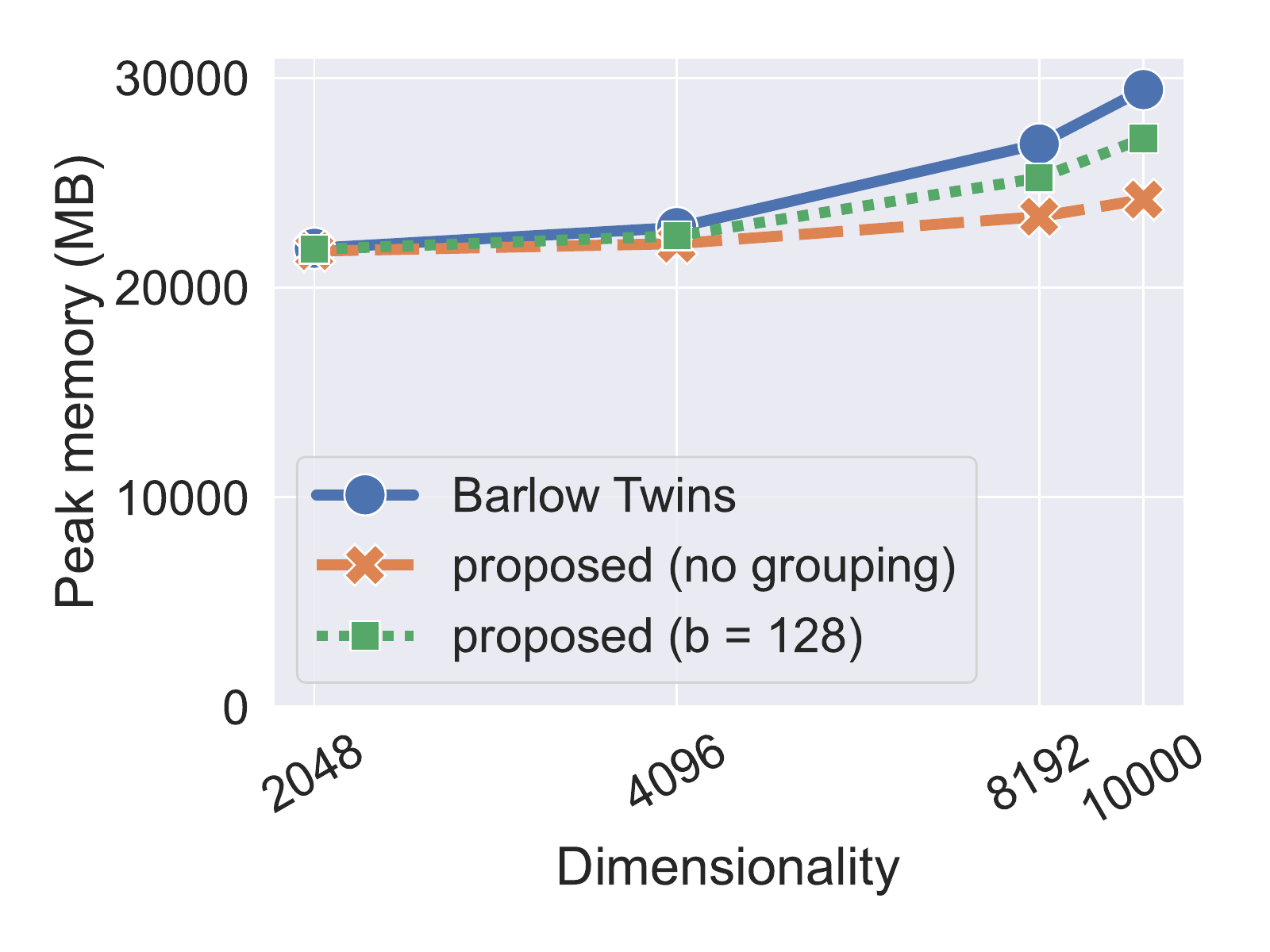}    & \includegraphics[scale=0.3]{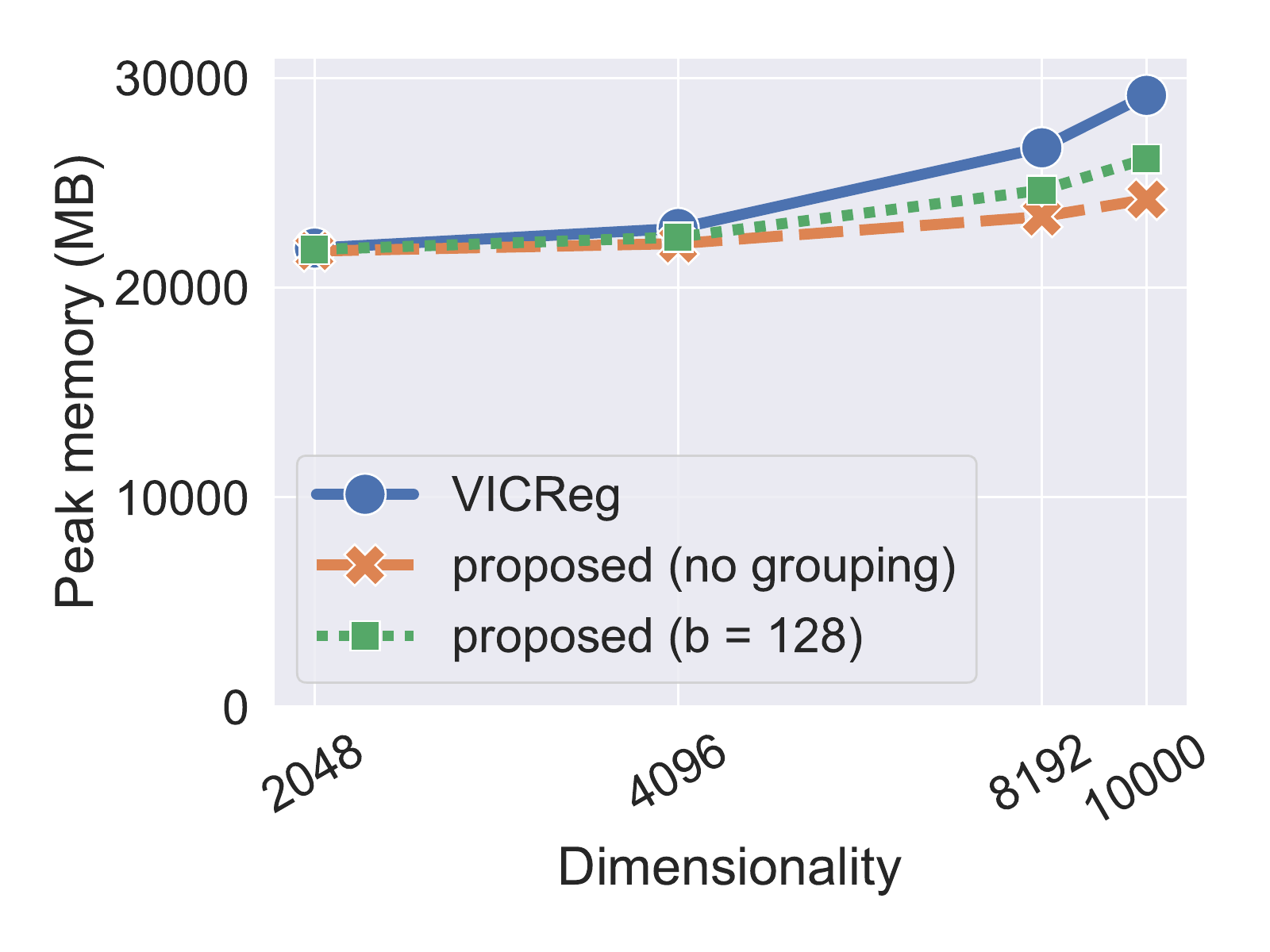}    \\[-1ex]
    \bottomrule
  \end{tabular}
  
  \caption{
    Training time and memory usage on ImageNet with ResNet-50 on a single GPU.
}
  \label{fig:dim-time-mem-imagenet-resnet50}
\end{figure}

Figure~\ref{fig:dim-time-mem-imagenet-resnet50} presents
the single GPU training cost with ResNet-50 on ImageNet.
Due to GPU memory limitation, we were unable to run Barlow Twins and VICReg at $d=16384$, and also the grouped version of the proposed models with block size $b=128$.
Instead of $d=16384$, we present the results at $d=10000$.\footnote{
  In Section~\ref{sec:time-ddp} (see Fig.~\ref{fig:dim-time-imagenet-resnet50-ddp-multinode}), we show the results with $d=16384$ using multi-node DDP computation.}
As in the results of ImageNet-100, we can observe that the proposed regularizers are more efficient than the existing regularizers.
At $d=8192$, the proposed model (without grouping) is
1.3 ($=41414.3/31280.0$)
times faster than VICReg, and
1.2 ($=37522.0/31306.0$)
times faster than Barlow Twins;
while at $d=10000$, it is 1.5 ($=47448.0/32208.3$)
times faster than VICReg, and
1.3 times ($=41546.3/32143.7$)
faster than Barlow Twins.

\subsection{Training Time with Distributed Data Parallel}
\label{sec:time-ddp}

\begin{figure}[tb]
  \centering
  \tablefont
  \begin{tabular}{l | m{.35\linewidth} | m{.35\linewidth}}
    \toprule
             & \multicolumn1{c|}{Cross-correlation regularization family}                                     & \multicolumn1{c}{Covariance regularization family}                                            \\
    \midrule                                                                                                                                                                                                   
    Elapsed time per 10 epochs & \includegraphics[scale=0.3]{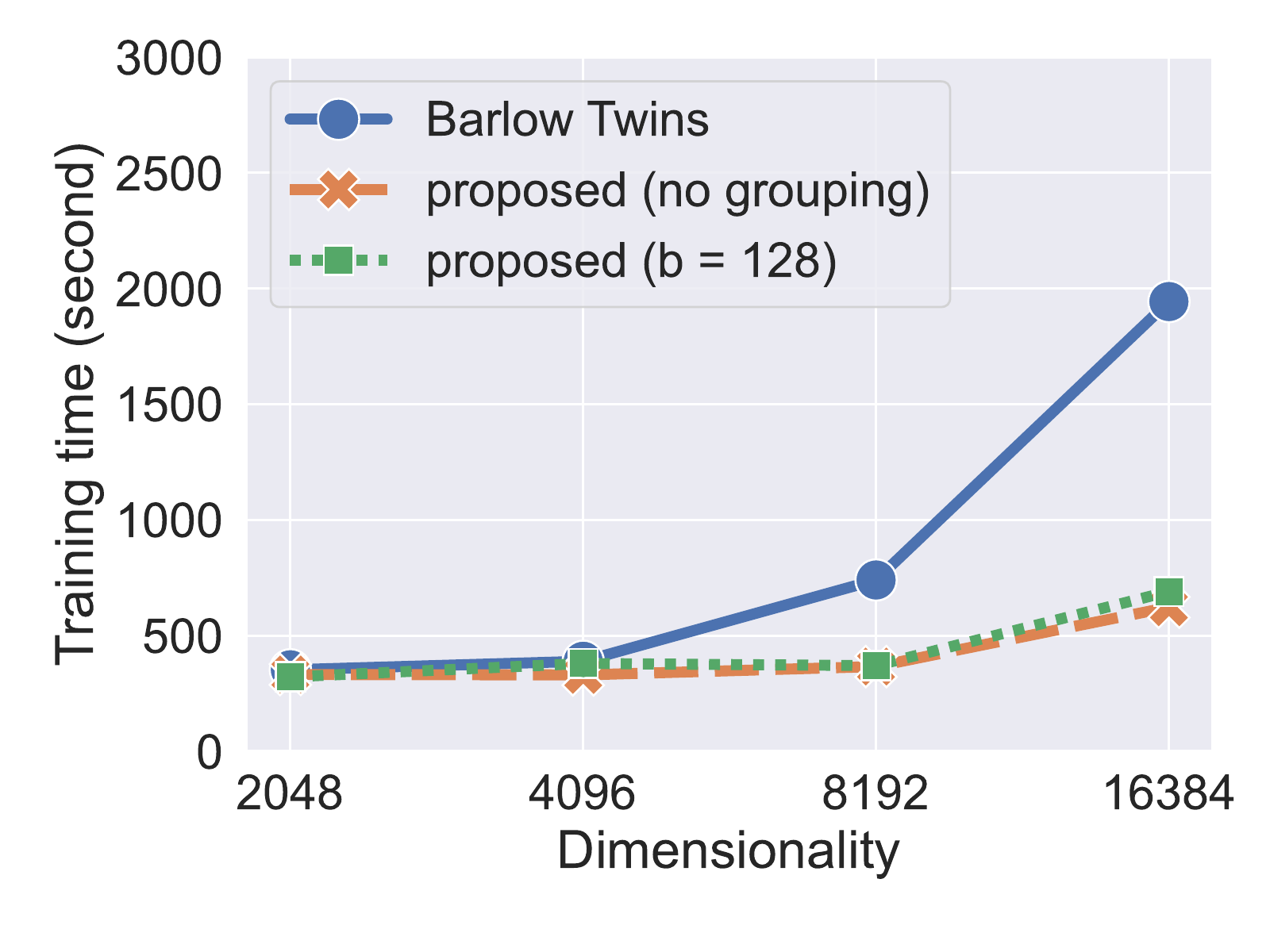} & \includegraphics[scale=0.3]{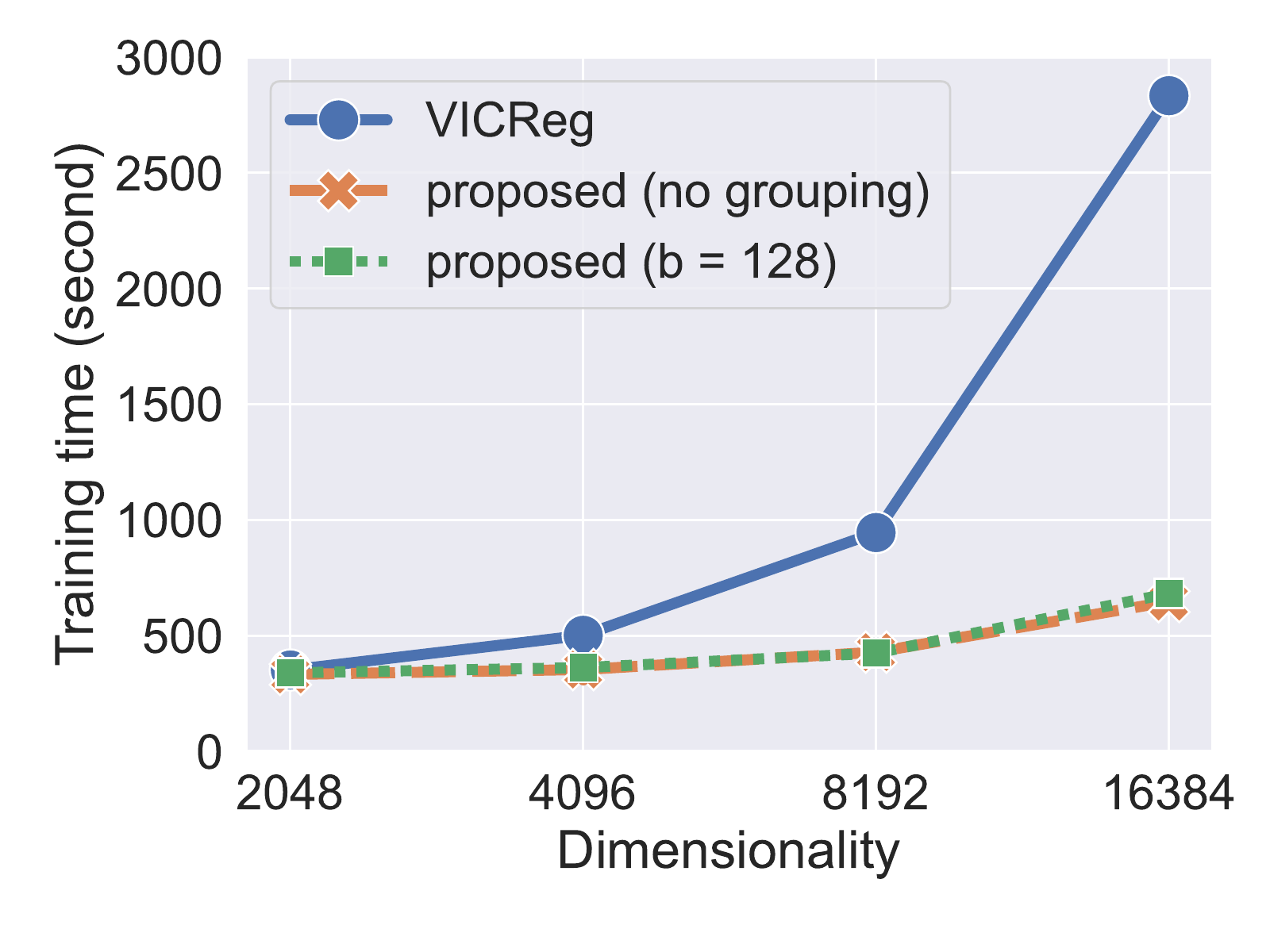} \\
    \bottomrule
  \end{tabular}

  \caption{
    The elapsed DDP training time on ImageNet-100 with ResNet-18.
}
  \label{fig:dim-time-imagenet100-resnet18-ddp}
\end{figure}

Except for \cref{tab:res-imagenet-time-gpu}, the training time figures reported in \cref{sec:experiment} and \crefappendix{sec:resnet50} were measured on a single GPU. Here we evaluate the timing of distributed data parallel (DDP) training, when multiple GPUs are available.
\cref{fig:dim-time-imagenet100-resnet18-ddp} shows the elapsed time of DDP training for ten epochs on eight A100 GPUs.
With DDP, the cost of communication between GPUs emerges as an additional factor determining the total computational time, 
and hence the relative merit of our method in reducing loss computation time is expected to diminish.
Although this is certainly true,
our method is still effective, improving the computation time by a factor of more than 2.2 ($=945.5/428.7$) for VICReg and 2.0 ($=740.6/366.4$) for Barlow Twins when $d=8192$ and a factor of 4.4 ($=2833.3/647.1$) and 3.1 ($=1943.7/622.7$) when $d=16384$.

\begin{figure}[tb]
  \centering
  \tablefont
  \begin{tabular}{l | m{.35\linewidth} | m{.35\linewidth}}
    \toprule
             & \multicolumn1{c|}{Cross-correlation regularization family}                                     & \multicolumn1{c}{Covariance regularization family}                                            \\
    \midrule                                                                                                                                                                                                   
    Elapsed time per 10 epochs & \includegraphics[scale=0.3]{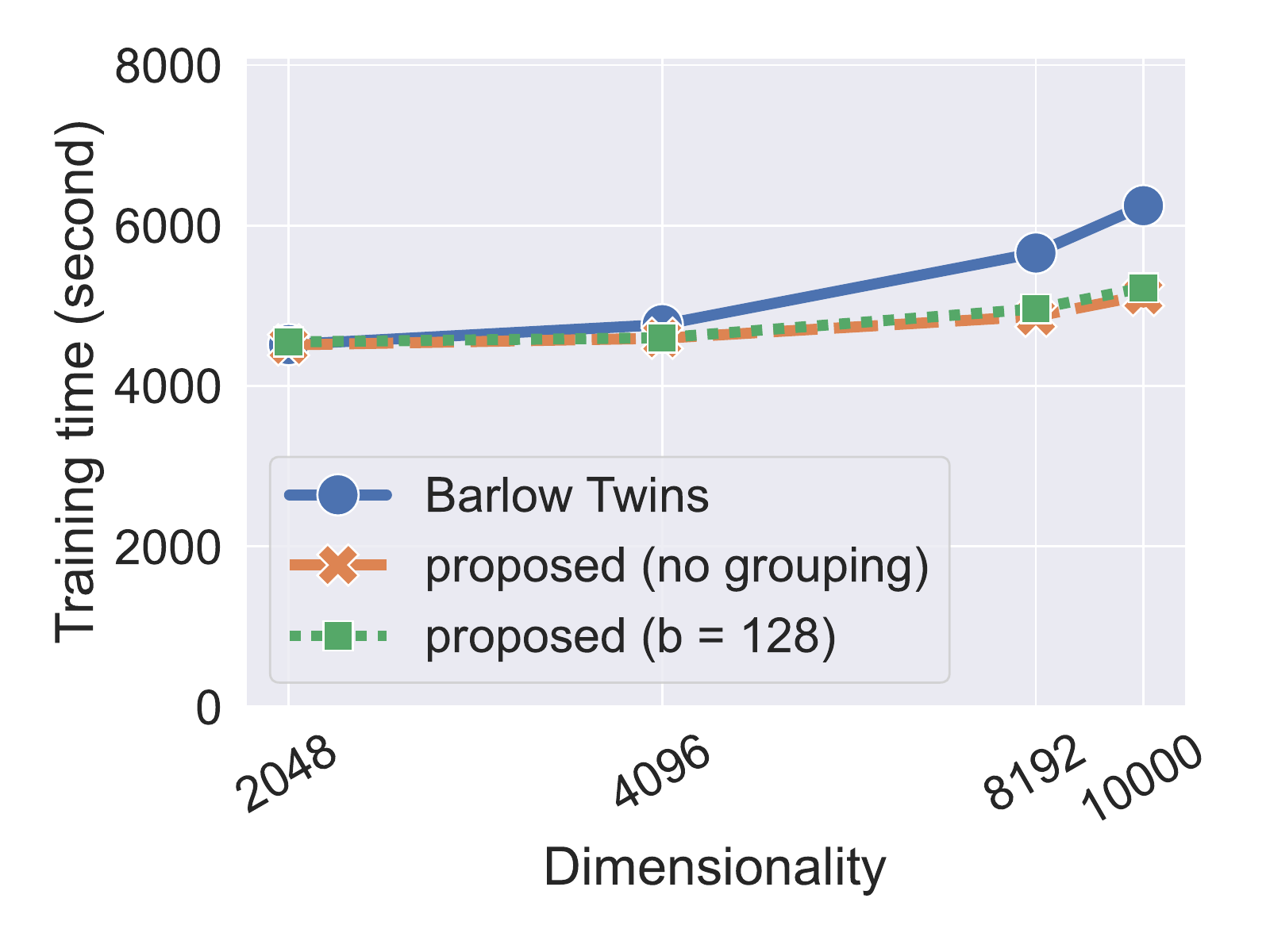} & \includegraphics[scale=0.3]{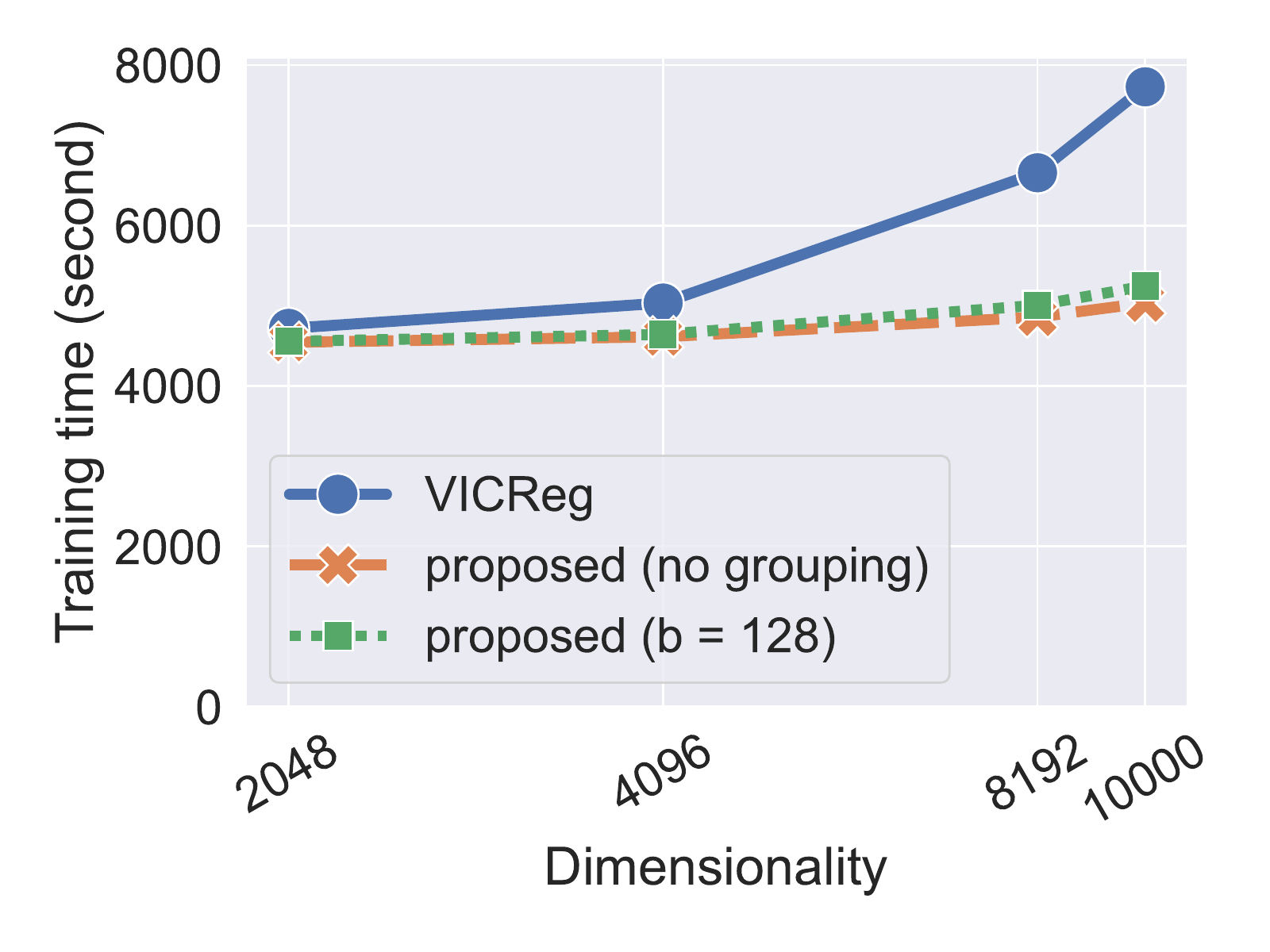} \\
    \bottomrule
  \end{tabular}

  \caption{
    The elapsed DDP training time on ImageNet with ResNet-50.
}
  \label{fig:dim-time-imagenet-resnet50-ddp}
\end{figure}

 \cref{fig:dim-time-imagenet-resnet50-ddp} shows the elapsed time on ImageNet with ResNet-50.
As in ImageNet-100 with ResNet-18 (\cref{fig:dim-time-imagenet100-resnet18-ddp}), our method improves speed, but the speed-up factor is smaller: 1.4 ($=6658.1/4858.9$) for VICReg and 1.2 ($=5657.6/4868.7$) for Barlow Twins when $d=8192$. 
 When $d=10000$, the factors are 1.5 ($=7729.9/5035.1$) for VICReg and 1.2 ($=6247.7/5129.6$) for Barlow Twins.

\begin{figure}[tb]
  \centering
  \tablefont
  \begin{tabular}{l | m{.35\linewidth} | m{.35\linewidth}}
    \toprule
             & \multicolumn1{c|}{Cross-correlation regularization family}                                     & \multicolumn1{c}{Covariance regularization family}                                            \\
    \midrule                                                                                                                                                                                                   
    Elapsed time per 10 epochs & \includegraphics[scale=0.3]{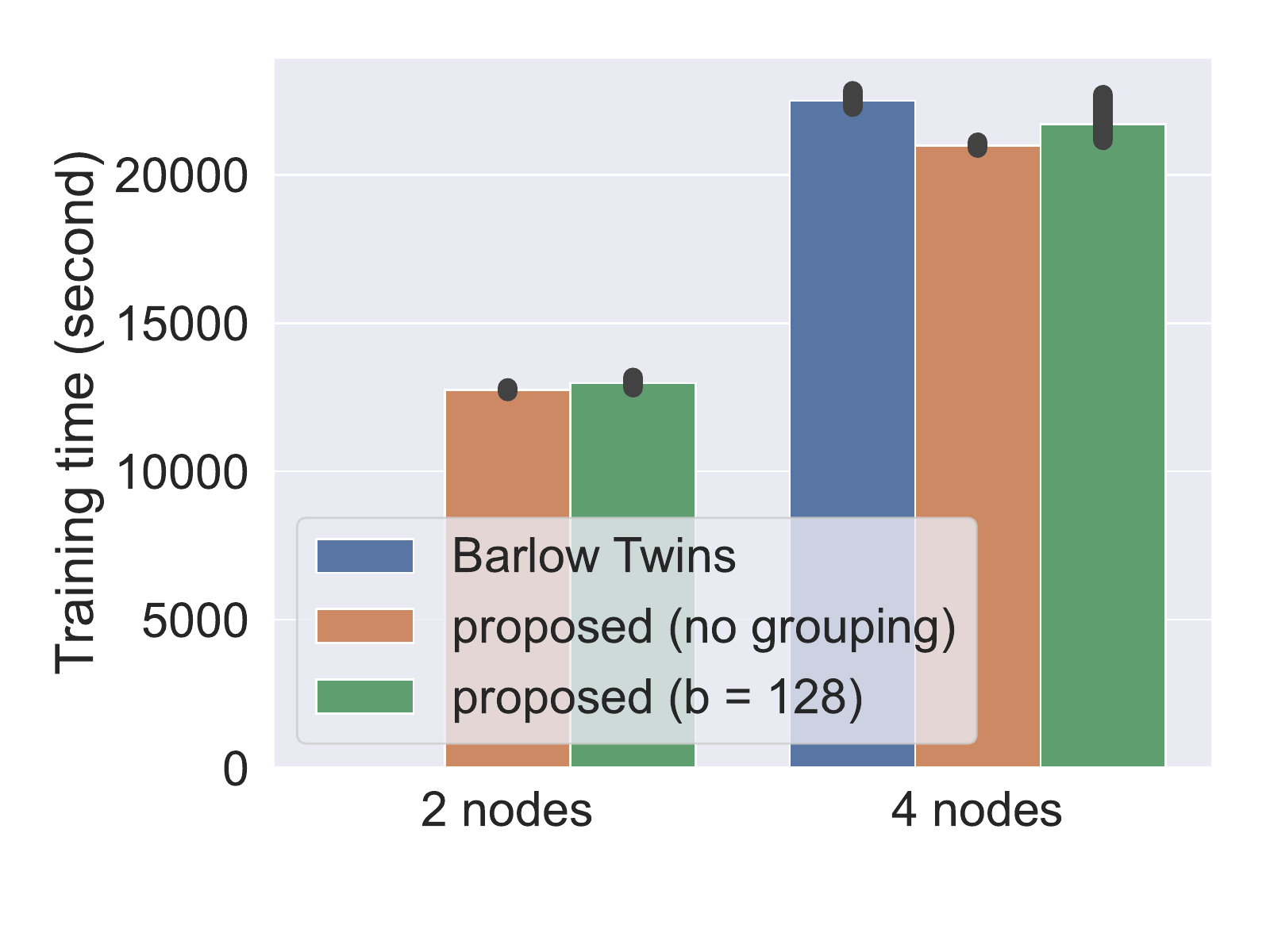} & \includegraphics[scale=0.3]{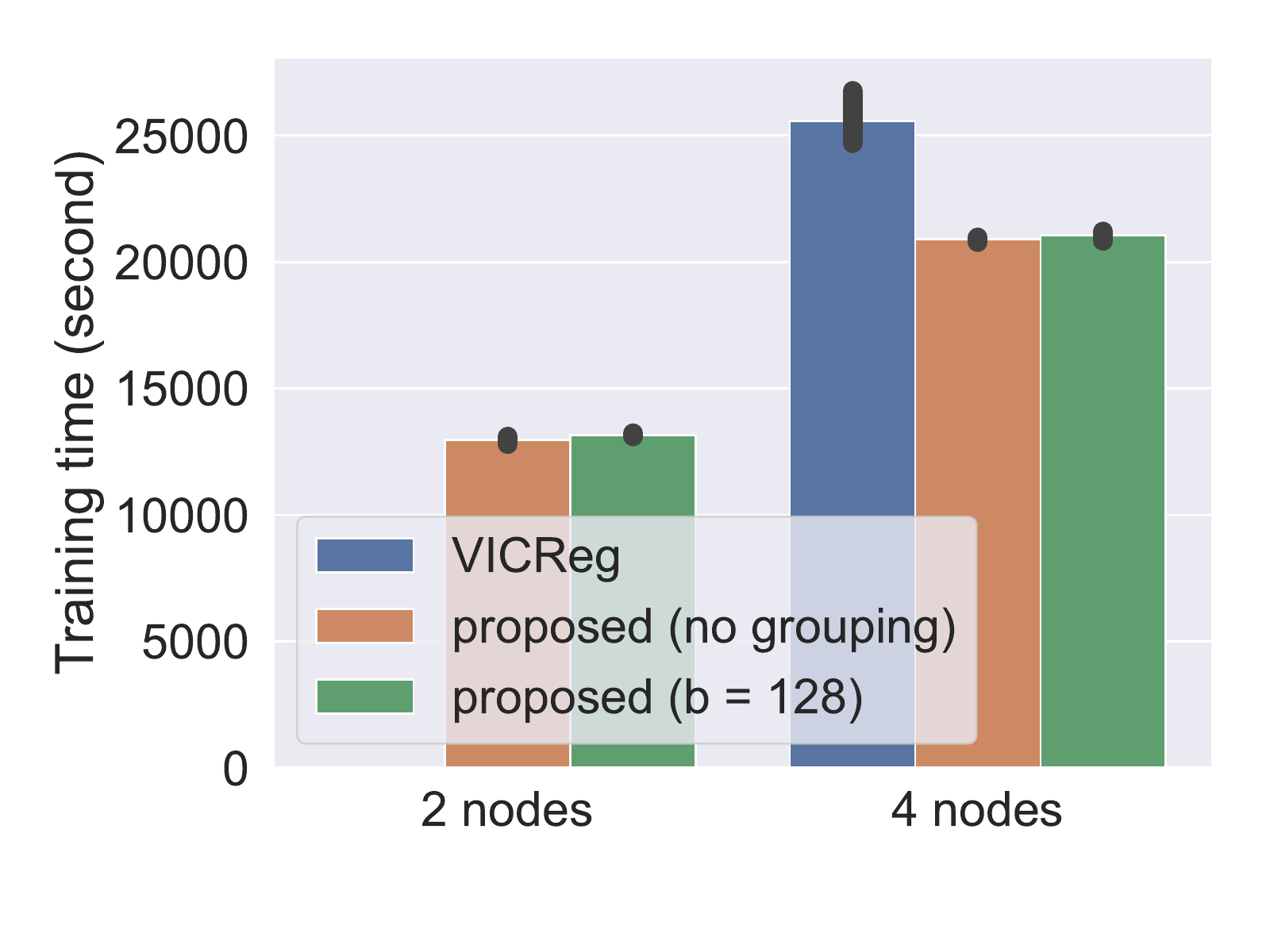} \\
    \bottomrule
  \end{tabular}
  
  \caption{
    The elapsed multi-node DDP training time on ImageNet with ResNet-50 ($d=16384$).
}
  \label{fig:dim-time-imagenet-resnet50-ddp-multinode}
\end{figure}

 In the ImageNet experiments (\cref{fig:dim-time-mem-imagenet-resnet50,fig:dim-time-imagenet-resnet50-ddp}), all models triggered an out-of-GPU-memory error when $d=16384$.
We thus use multi-node DDP training for $d=16384$.
We evaluated two situations: training using 2 nodes and 4 nodes.
In both situations, we set the effective batch size to 1024.
Figure~\ref{fig:dim-time-imagenet-resnet50-ddp-multinode} shows the results.
Barlow Twins and VICReg still ran out of memory under 2 nodes, but the proposed models (with or without grouping) worked in this situation
thanks to their efficient memory usage. 
With 4 nodes, all models were trained successfully, and the proposed models were slightly faster than Barlow Twins and VICReg.
However,
in this setting, there is no point in training our models using 4 nodes, when they are trainable on 2 nodes.
If we compare the speed of our models trained on 2 nodes with Barlow Twins and VICReg (which failed to be trained on 2 nodes) on 4 nodes, the advantage of our models becomes more pronounced.

\subsection{Computation Time for Forward and Backward Passes}
\label{sec:time-analysis}

We analyzed the training time spent for the forward pass (in the model network excluding the loss node), forward loss computation, and backpropagation.
These three types are denoted by ``Forward (model)'', ``Forward (loss)'', and ``Backward (model + loss)'', respectively.
These measurements were quoted from the ``Total time (s)'' column in the respective lines in the output of the Simple Profiler in PyTorch Lightning.
Note that by default, the profiler only reports total forward computation time.
To separate the total time into ``Forward (model)'' and ``Forward (loss)'', we specified custom measurement points in our code where the model (backbone network and projection network) forward computation and the loss computation are performed.\footnote{\url{https://pytorch-lightning.readthedocs.io/en/1.6.4/advanced/profiler.html\#profile-logic-of-your-interest}}
Since the custom measurement points did not work for backpropagation, we plot the value of ``Strategy.backward'' line as the total backpropagation time ``Backward (model + loss)''.

\cref{fig:dim-time-all-bar} presents the results.
The times for loss computation and backpropagation both improved with our method;
see \cref{tab:res-imporovement-forward-backward-imagenet100-resnet18,tab:res-imporovement-forward-backward-imagenet-resnet50} for the detail of improvements.
The improvement in backpropagation is due to reduced time at the loss node, because Barlow Twins and the proposed method have the same model network and differ only in the loss used.
This reduction in the backpropagation time is not surprising because the same computation graph is traversed in both the forward and backward passes, and therefore their asymptotic computational complexity is roughly identical.

\begin{figure}[htb]
  \centering
  \tablefont
  \begin{tabular}{ >{\centering} m{.12\linewidth} |  m{.33\linewidth} | m{.33\linewidth}}
    \toprule
    Setting                                                                                     & \multicolumn1{c|}{Cross-correlation regularization}                                           & \multicolumn1{c}{Covariance regularization}                                                   \\
    \midrule                                                                                                                                                                                                   
    (a)\par ImageNet-100\par ResNet-18\par 1 GPU                                                & \includegraphics[scale=0.3]{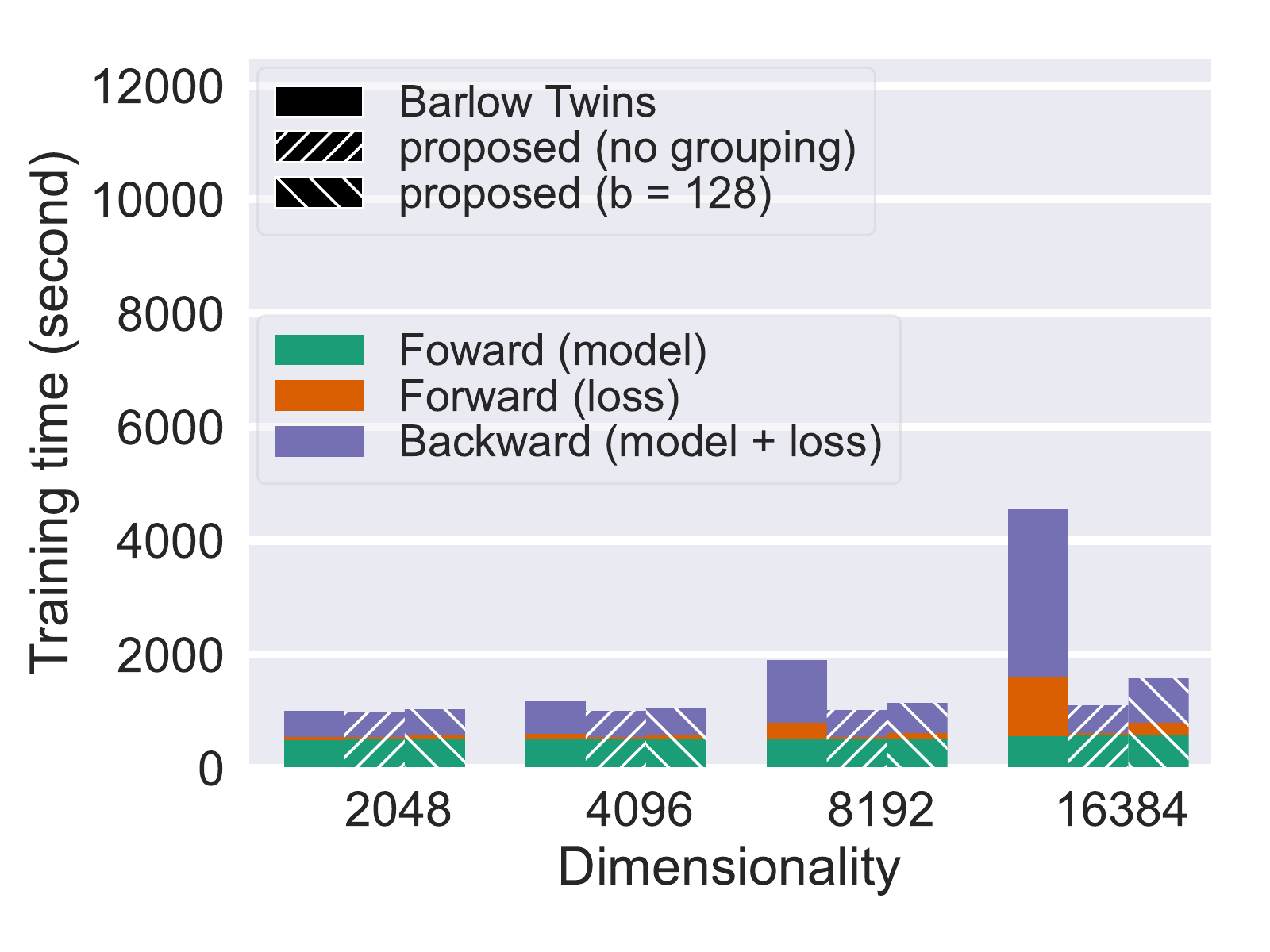} & \includegraphics[scale=0.3]{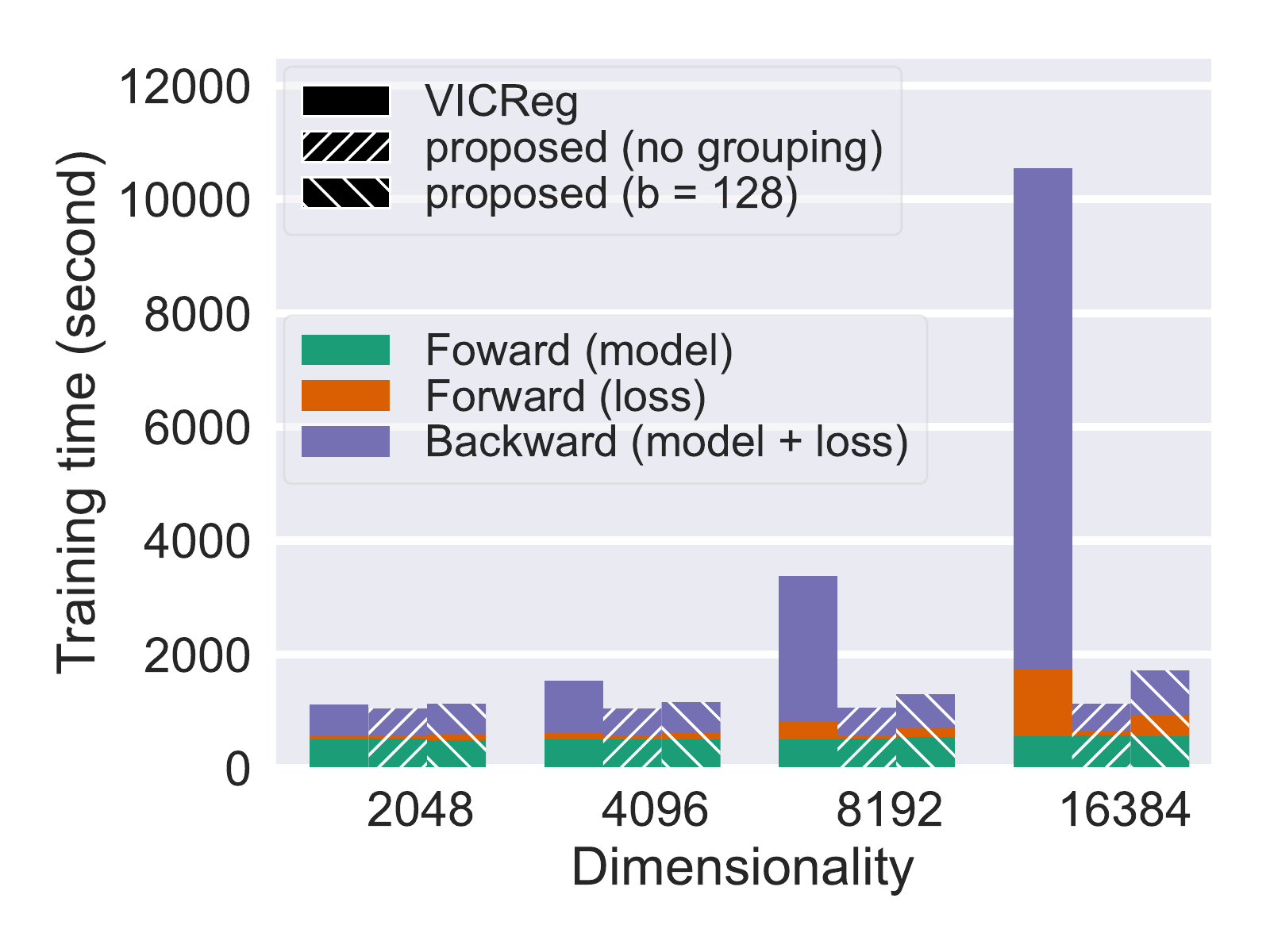} \\
    \midrule
    (b)\par ImageNet\par ResNet-50\par 1 GPU                                                    & \includegraphics[scale=0.3]{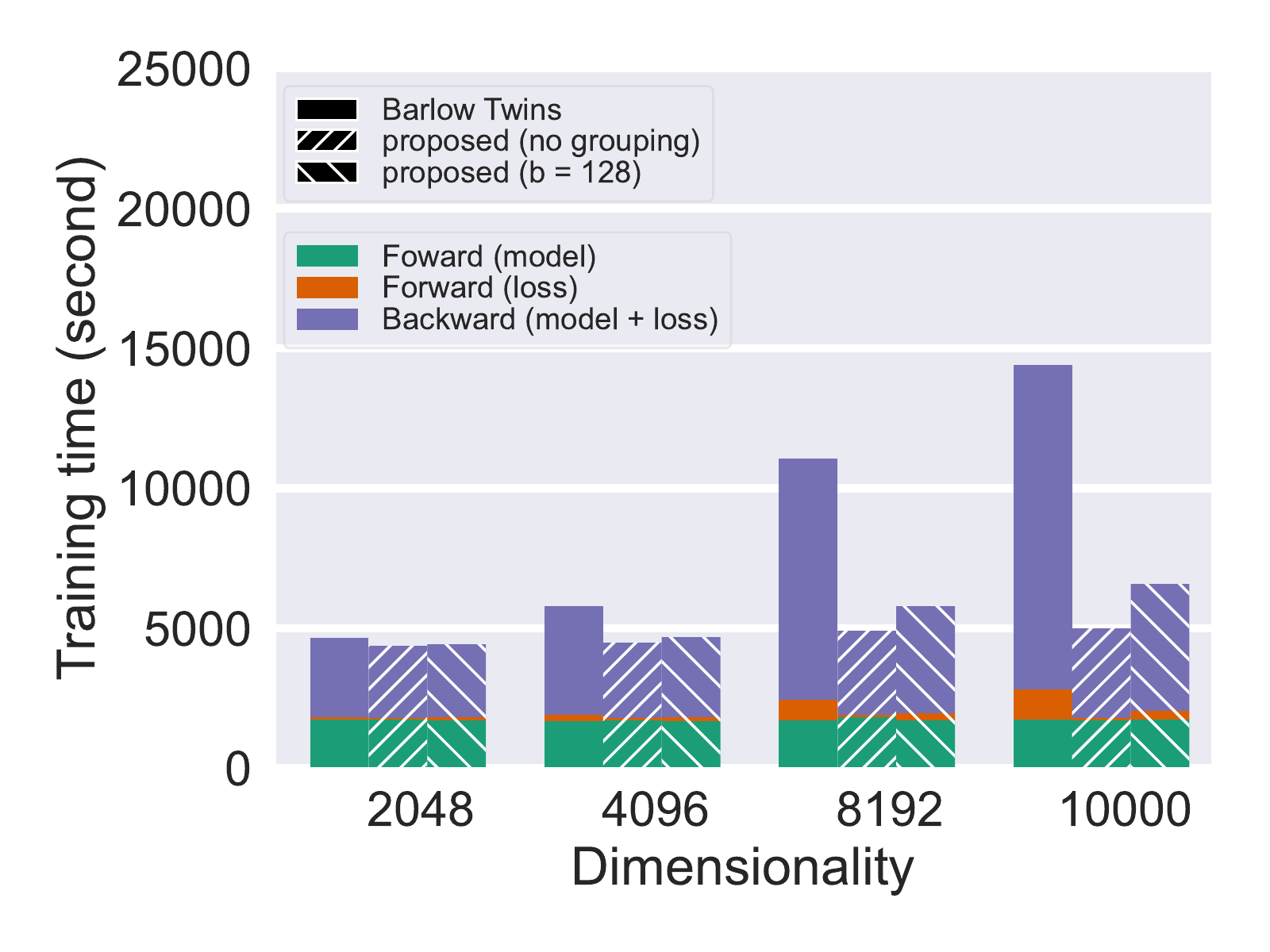}    & \includegraphics[scale=0.3]{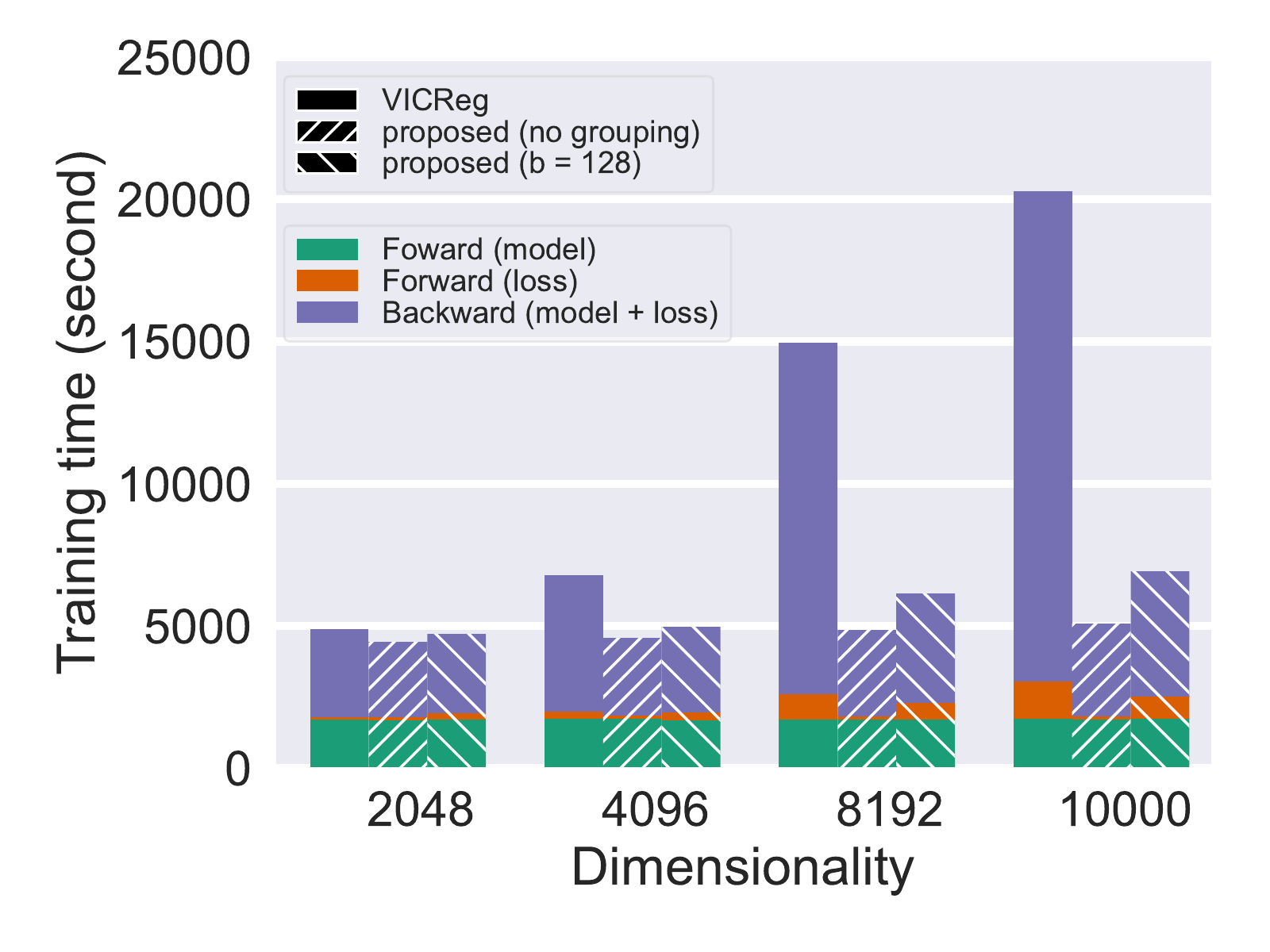}    \\
    \midrule
    (c)\par ImageNet-100\par ResNet-18\par 8-GPU DDP
                                                                                                & \includegraphics[scale=0.3]{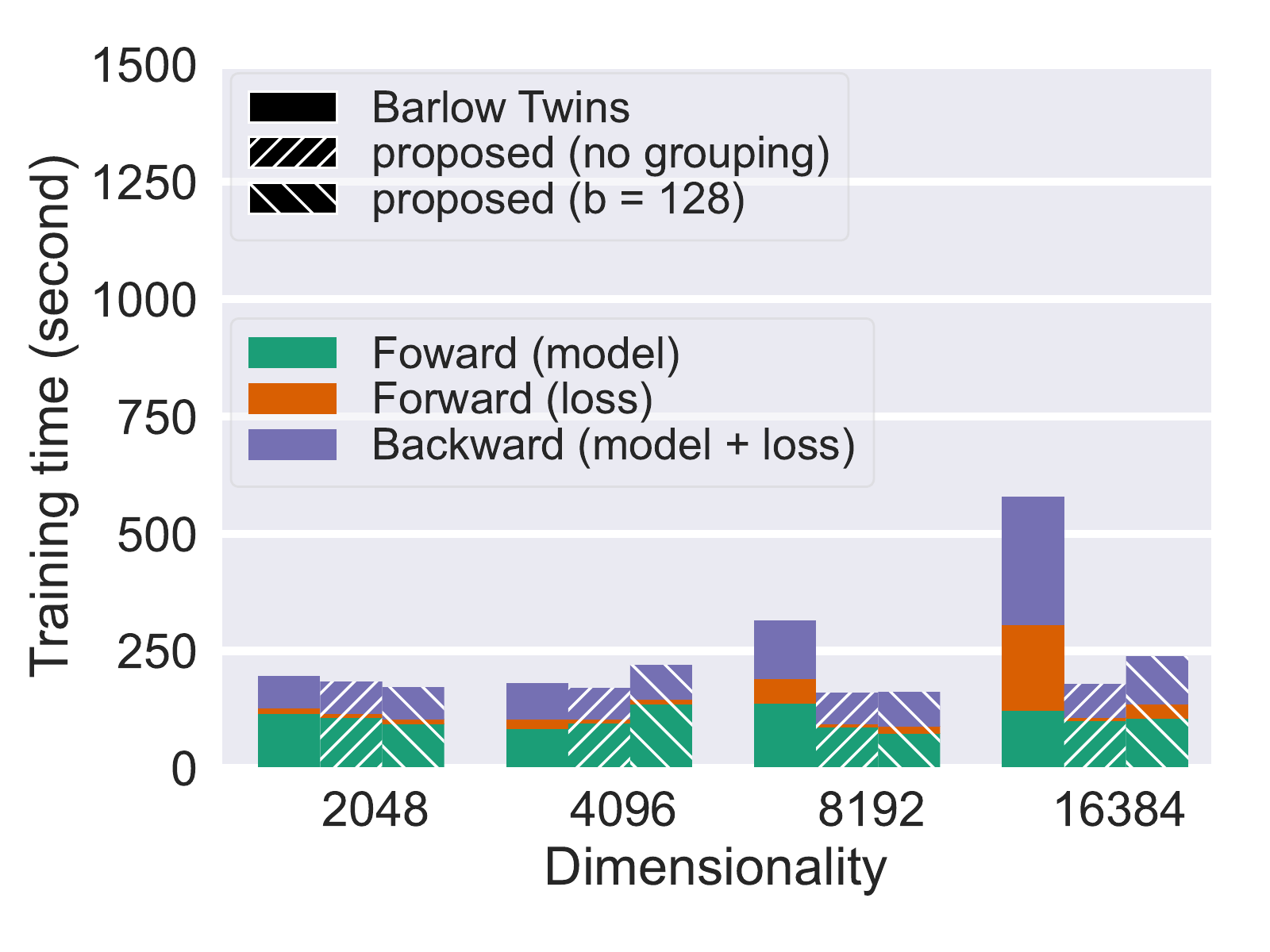}  & \includegraphics[scale=0.3]{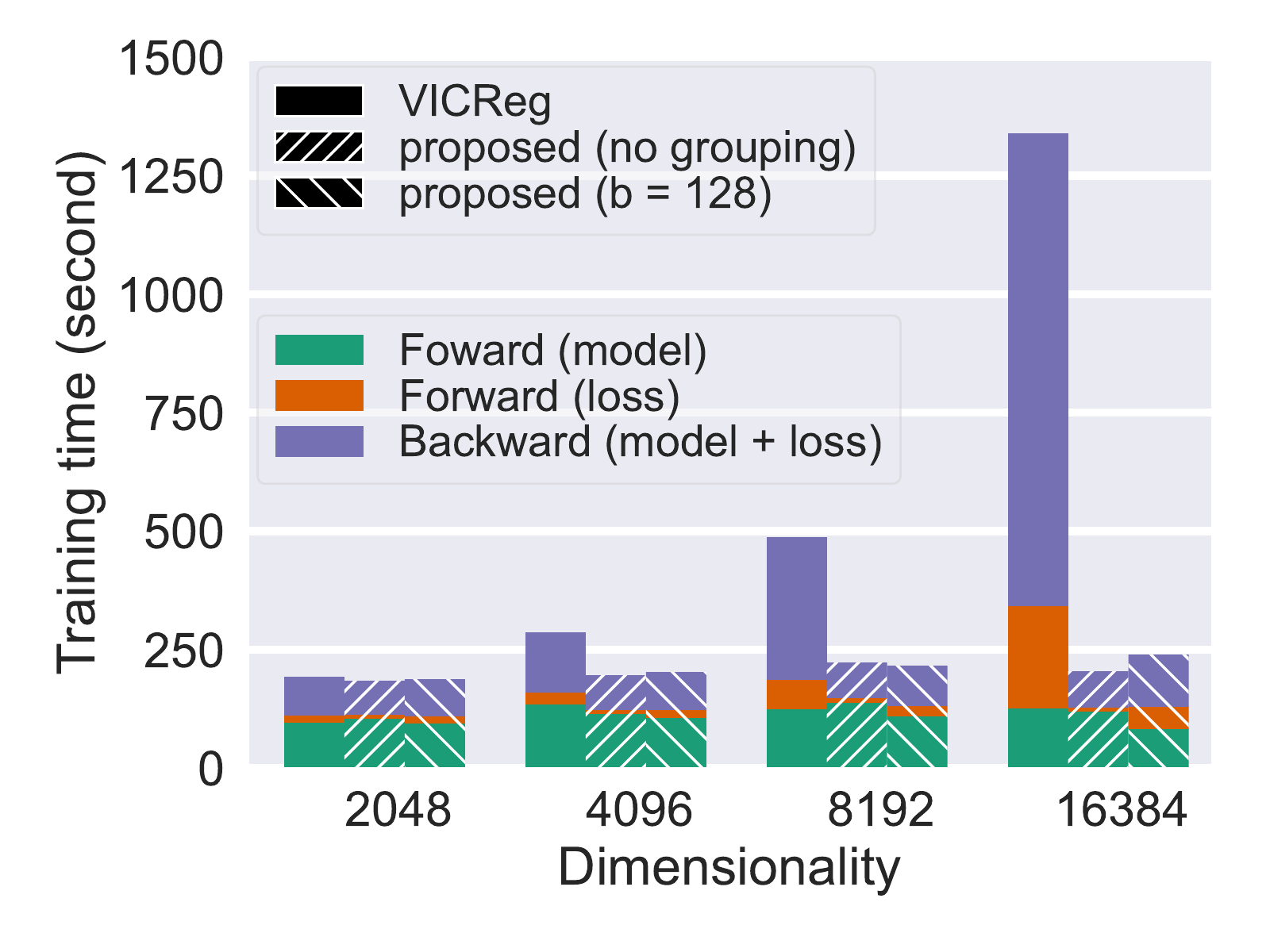}  \\
    \midrule
    (d)\par ImageNet\par ResNet-50\par 8-GPU DDP                                                & 
      \includegraphics[scale=0.3]{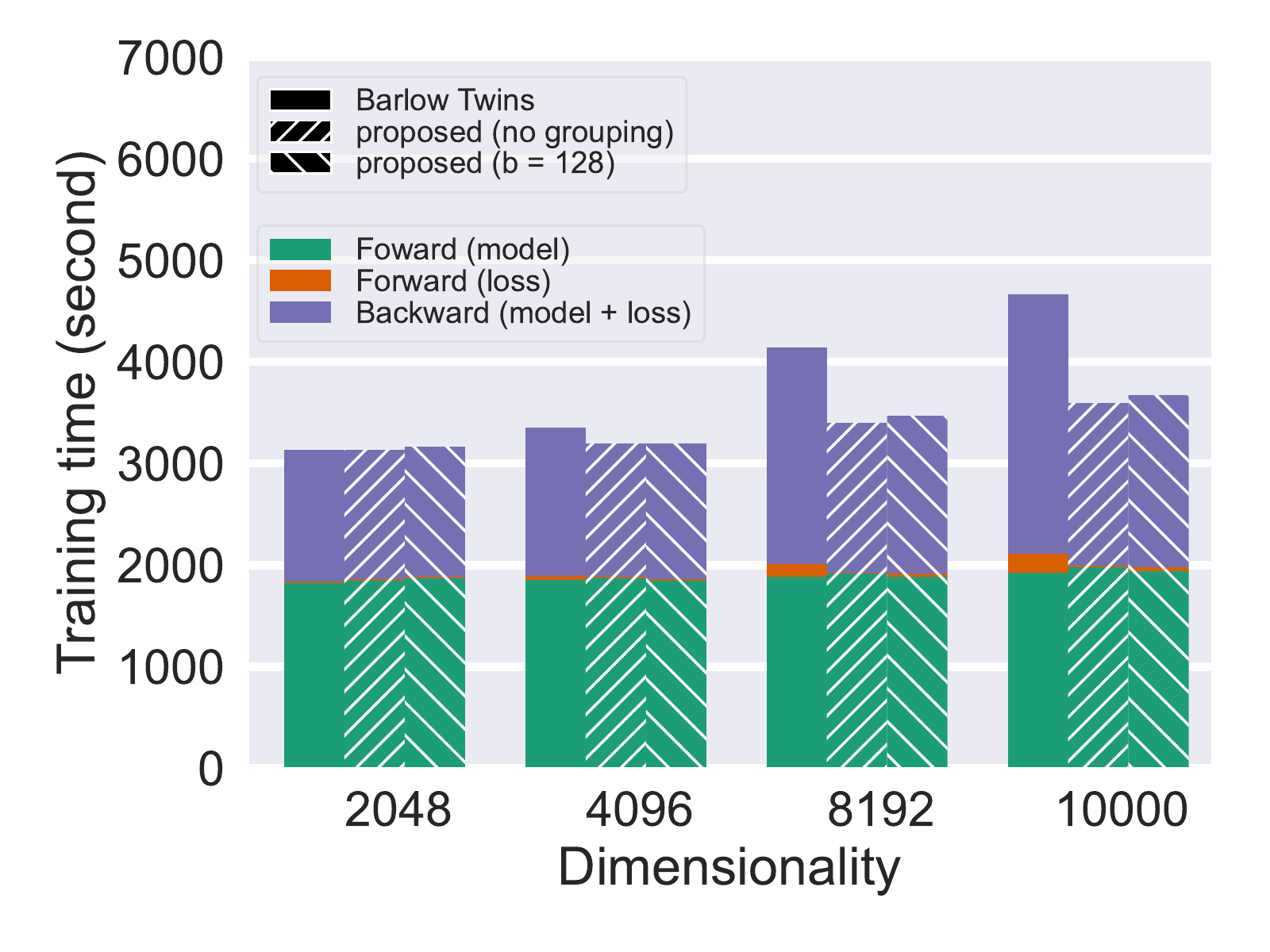} & \includegraphics[scale=0.3]{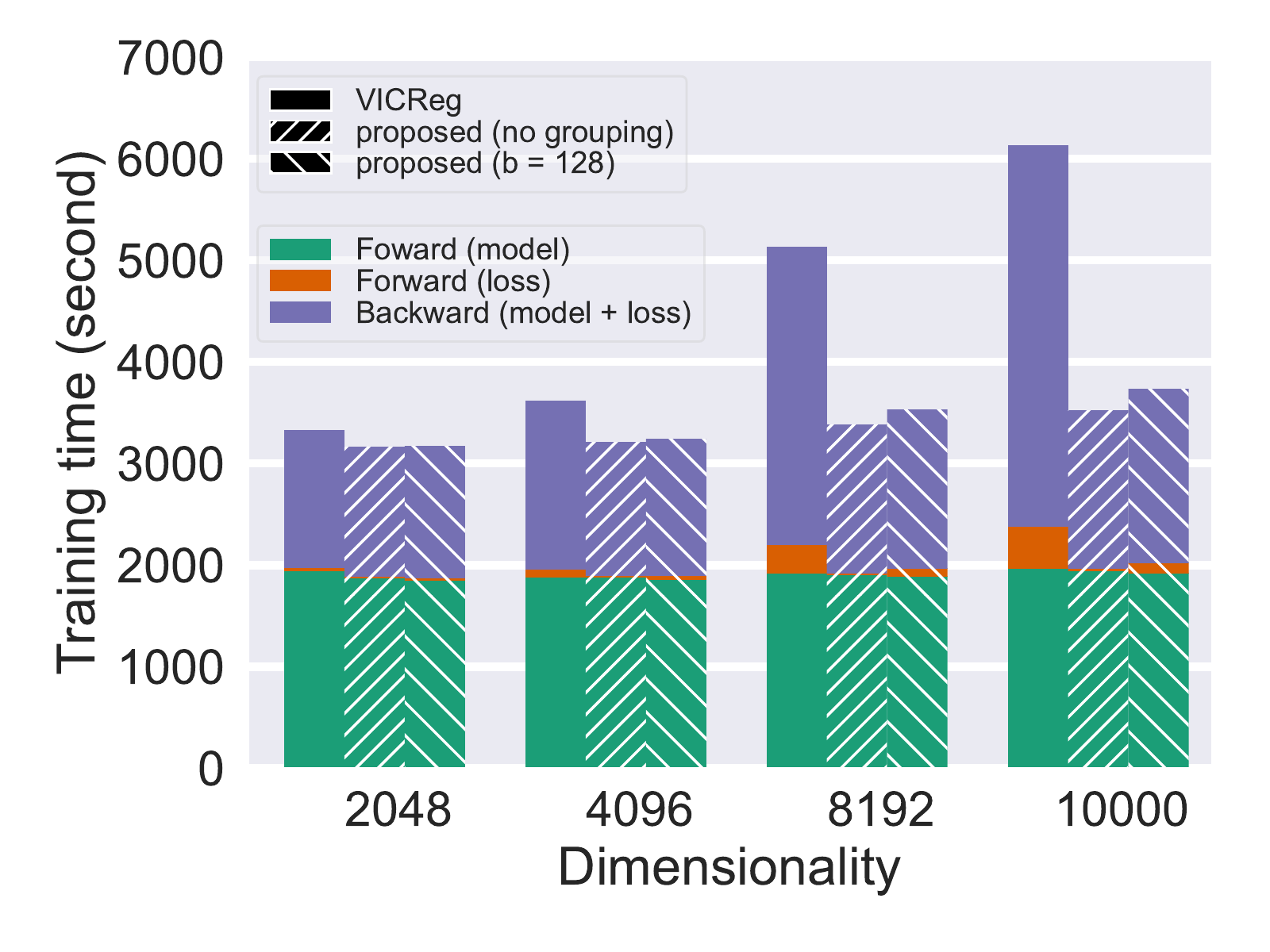}                                                                                                     \\
      \bottomrule
    \end{tabular}

    \caption{The elapsed 10-epoch training time spent on forward model (green) and loss (orange) computation, and backpropagation (total of model and loss; purple).}
    \label{fig:dim-time-all-bar}
\end{figure}

\begin{table}[tb]
  \caption{
    The improvements for loss and backward computations on ImageNet100 with ResNet-18. 
  }
  \label{tab:res-imporovement-forward-backward-imagenet100-resnet18}
  \centering
  \tablefont
  \begin{tabular}{lcr ll}
    \toprule 
    Model               & \#GPU & $d$   & Forward (loss)                 & Backward (model + loss)                   \\
    \midrule
    Barlow Twins--style & 1     & 8192  & \phantom{0}7.5 ($=285.5/38.0$) & \phantom{0}2.5 ($=1101.6/448.6$)          \\
                        &       & 16384 & \textbf{23.1} ($=1038.7/44.9$) & \phantom{0}\textbf{6.3} ($=2959.1/468.0$) \\
    \cmidrule{2-5}
                        & 8     & 8192  & \phantom{0}7.4 ($=52.5/7.1$)   & \phantom{0}1.9 ($=124.3/67.1$)            \\
                        &       & 16384 & \textbf{24.8} ($=183.6/7.4$)   & \phantom{0}\textbf{3.9} ($=274.0/70.3$)   \\
    \midrule
    VICReg-style        & 1     & 8192  & \phantom{0}6.0 ($=323.0/53.5$) & \phantom{0}5.4 ($=2548.8/473.2$)          \\
                        &       & 16384 & \textbf{19.5} ($=1166.5/59.8$) & \textbf{18.3} ($=8820.3/482.0$)           \\
    \cmidrule{2-5}
                        & 8     & 8192  & \phantom{0}7.0 ($=62.6/8.9$)   & \phantom{0}4.1 ($=300.5/74.2$)            \\   
                        &       & 16384 & \textbf{24.2} ($=215.7/8.9$)   & \textbf{13.2} ($=996.3/75.5$)             \\
    \bottomrule
  \end{tabular}
\end{table}

\begin{table}[tb]
  \caption{
    The improvements for loss and backward computations on ImageNet with ResNet-50. 
  }
  \label{tab:res-imporovement-forward-backward-imagenet-resnet50}
  \centering
  \tablefont
  \begin{tabular}{lcr ll}
    \toprule 
    Model               & \#GPU & $d$   & Forward (loss)                  & Backward (model + loss)          \\
    \midrule
    Barlow Twins--style & 1     & 8192  & \phantom{0}9.5 ($=751.1/79.2$)  & 2.9 ($=8594.0/2977.5$)           \\
                        &       & 10000 & \textbf{13.2} ($=1097.5/83.2$)  & \textbf{3.6} ($=11583.0/3181.2$) \\
    \cmidrule{2-5}
                        & 8     & 8192  & 10.6 ($=125.8/11.9$)            & 1.5 ($=2129.1/1463.5$)           \\
                        &       & 10000 & \textbf{15.0} ($=184.4/12.3$)   & \textbf{1.6} ($=2552.2/1593.5$)  \\
    \midrule
    VICReg-style        & 1     & 8192  & \phantom{0}8.4 ($=898.8/107.0$) & 4.1 ($=12346.0/3016.6$)          \\
                        &       & 10000 & \textbf{11.9} ($=1311.1/110.3$) & \textbf{5.4} ($=17249.3/3221.7$) \\
    \cmidrule{2-5}
                        & 8     & 8192  & 18.1 ($=280.7/15.5$)            & 2.0 ($=2937.3/1467.1$)           \\   
                        &       & 10000 & \textbf{25.2} ($=415.7/16.5$)   & \textbf{2.4} ($=3752.4/1557.9$)  \\
    \bottomrule
  \end{tabular}
\end{table}

\FloatBarrier

\section{Code}
\label{sec:pseudocode}

\cref{lst:covsum,lst:xcorr} present Python-based implementations for covariance and cross-correlation regularizers (without feature grouping). 
As explained in \cref{sec:method}, the summary vectors can be efficiently calculated with FFT (see Listing~\ref{lst:sumvec}).

In the computation of the proposed regularizers, we do not conduct collective operations, such as all-reduce and gather functions.

\begin{lstlisting}[float=tbp, language=Python, caption=Computing Barlow Twins--style cross-correlation regularizer, label=lst:xcorr]
# Z1, Z2: projected image embeddings (n x d)
# q: a hyperprameter for L_q^q norm

def xsum_regularizer(Z1, Z2, G, q):
    # pre-process: centering and normalization
    Z1 = batch_normalization(Z1)
    Z2 = batch_normalization(Z2)
 
    # feature permutation
    idx = torch.randperm(Z1.shape[1])
    Z1 = Z1[:, idx]
    Z2 = Z2[:, idx]

    # summary vector
    sumvec = cal_sumvec(Z1, Z2, 0) / n

    # loss for off-diagonal elements
    if q == 1:
        loss = torch.sum(sumvec[1:].abs())
    elif q == 2:
        loss = torch.sum(sumvec[1:].pow(2))

    return loss
\end{lstlisting}
 \begin{lstlisting}[float=tbp, language=Python, caption=Computing VICReg-style covariance regularizer, label=lst:covsum]
# Z: projected image embeddings ([n: batch size] x [d: embedding dimension])
# q: a hyperparameter for L_q^q norm

def covsum_regularizer(Z, blck_size, q):
    # pre-process: centering
    Z =  Z - Z.mean(dim=0)
    
    # feature permutation
    idx = torch.randperm(Z.shape[1])
    Z = Z[:, idx] 
    
    # summary vector
    sumvec = cal_sumvec(Z, Z, 0) / (n - 1)
    
    # loss for off-diagonal elements
    if q == 1:
        loss = torch.sum(sumvec[1:].abs())
    elif q == 2:
        loss = torch.sum(sumvec[1:].pow(2))
    
    return loss
\end{lstlisting}
 \begin{lstlisting}[float=tbp,language=Python, caption=Summary vector computation, label=lst:sumvec]
def cal_sumvec(z1, z2, dim):
  fz1 = fft.rfft(z1)
  fz2 = fft.rfft(z2)
  fz1_conj = fz1.conj()
  fz_prod = fz1_conj * fz2
  fc = fz_prod.sum(dim=dim)
  sumvec= fft.irfft(fc)
  return sumvec
\end{lstlisting}

\bibliographystyleappendix{alpha} \bibliographyappendix{ref}

\fi

\end{document}